\documentclass[smallcondensed,final]{article}     
\usepackage{authblk} 

\usepackage{graphicx}
\usepackage{amsmath, amssymb, amsfonts, bm}
\usepackage{ulem}
\usepackage[ruled, vlined]{algorithm2e}
\usepackage{algorithmic}
\usepackage{xcolor} 
\usepackage{color}
\usepackage[colorlinks,linkcolor=blue,citecolor=blue, anchorcolor = blue]{hyperref}
\usepackage{natbib}
\usepackage[labelfont=bf, labelsep=space]{subcaption}
\usepackage[labelfont=bf, labelsep=space]{caption}
\usepackage{booktabs}   
\usepackage{multirow} 
\usepackage{scalefnt}
\usepackage{colortbl}   
\usepackage{diagbox}
\usepackage{appendix}
\usepackage[left=2cm,right=2cm,top=2cm,bottom=2cm]{geometry}
\begin{document}

\title{Partially  Hidden  Markov  Chain  Linear  Autoregressive  model:\\  inference and forecasting
}

\makeatletter
\newcommand\email[2][]%
   {\newaffiltrue\let\AB@blk@and\AB@pand
      \if\relax#1\relax\def\AB@note{\AB@thenote}\else\def\AB@note{\relax}%
        \setcounter{Maxaffil}{0}\fi
      \begingroup
        \let\protect\@unexpandable@protect
        \def\thanks{\protect\thanks}\def\footnote{\protect\footnote}%
        \@temptokena=\expandafter{\AB@authors}%
        {\def\\{\protect\\\protect\Affilfont}\xdef\AB@temp{#2}}%
         \xdef\AB@authors{\the\@temptokena\AB@las\AB@au@str
         \protect\\[\affilsep]\protect\Affilfont\AB@temp}%
         \gdef\AB@las{}\gdef\AB@au@str{}%
        {\def\\{, \ignorespaces}\xdef\AB@temp{#2}}%
        \@temptokena=\expandafter{\AB@affillist}%
        \xdef\AB@affillist{\the\@temptokena \AB@affilsep
          \AB@affilnote{}\protect\Affilfont\AB@temp}%
      \endgroup
       \let\AB@affilsep\AB@affilsepx
}
\makeatother

\author[$\dag$]{Fatoumata Dama}
\author[$\dag$]{Christine Sinoquet}
\affil[$\dag$]{LS2N / UMR CNRS 6004, Nantes University, France}
\email{\url{{fatoumata.dama,christine.sinoquet}@univ-nantes.fr}} 

\date{}

\maketitle

\begin{abstract}
Time series subject to change in regime have attracted much interest in domains such as econometry, finance or meteorology. For discrete-valued regimes, some models such as the popular Hidden Markov Chain (HMC) describe time series whose state process is {\it unknown} at all time-steps. 
Sometimes, time series are firstly labelled thanks to some annotation function. Thus, another category of models handles the case with regimes {\it observed} at all time-steps. We present a novel model which addresses the intermediate case: (i) state processes associated to such time series are modelled by Partially Hidden Markov Chains (PHMCs); (ii) a linear autoregressive (LAR) model drives the dynamics of the time series, within each regime. We describe a variant of the expection maximization (EM) algorithm devoted to PHMC-LAR model learning. We propose a hidden state inference procedure and a forecasting function that take into account the observed states when existing. We assess inference and prediction performances, and analyze EM convergence times for the new model, using simulated data. We show the benefits of using partially observed states to decrease EM convergence times. A fully labelled scheme with unreliable labels also speeds up EM. This offers promising prospects to enhance PHMC-LAR model selection. We also point out the robustness of PHMC-LAR to labelling errors in inference task, when large training datasets and moderate labelling error rates are considered. Finally, we highlight the remarkable robustness to error labelling in the prediction task, over the whole range of error rates.
\end{abstract}


{\bf Keywords} Time series analysis . Autoregressive model . Regime-switching model . Markov chain . \\
\verb^             ^ \vspace{3mm}Forecasting . Hidden state inference


\section{Introduction}
\label{intro}
Time series are widely present in many domains such as industry, energy, meteorology, e-commerce, social networks or health. They represent the temporal evolving of systems and help us to understand their temporal dynamics and perform short-, medium- or long-term predictions. A major research line has been dedicated to time series analysis. In this line, exponential smoothing models \citep{gardner_2006_int-journ-forecasting_exponent-smooth-paramet-initial-val-select, bergmeir_hyndman_benitez_2016_intern-journ-of-forecasting_expon-smooth-stl-plus-box-cox}, Box and Jenkins models \citep{box_jenkins_reinsel_et_al_2015_book_time_series_analysis} and nonlinear autoregressive neural networks
\citep{yu_zhou_tan_2014_jour-plos-one_ARIMA-and-NARNN, wang_wang_yan_2019_journ-ieee_NARNN, noman_alkawsi_alkahtani_2020_journ-elsevier_NARNN-wind-time-series} are essentially devoted to forecasting. In addition to the forecasting goal, regime-switching autoregressive models \citep{ubilava_helmesr_2013_environ-modelling-softwa_smooth-transition-autoregres-model, hamilton_1990_journ-econometrics_time_series-regime-changes} also allow to discover hidden behaviors of such systems.

In the cases when the studied system is \textbf{stationary}, that is its behavior is time-independent, the Linear AutoRegressive (LAR) model is a framework widely used to capture the autoregressive dynamics of the corresponding time series \citep{wold_1954_book_stationary-time-series, degtyarev_Gankevich_2019_book_AR-model-of-irregular-waves}. The LAR model is a simple linear regression model in which predictors are lagged values of the current value in the time series. However, many real-life systems are subject to changes in behaviors: for instance in econometry, we distinguish between recession and expansionary phases; in meteorology, anticyclonic conditions alternate with low pressure conditions. These systems are commonly referred to as \textbf{regime-switching systems}, where each regime corresponds to a specific behavior. Each time-step is associated with some state, amongst those allowed for the system. Regime-switching system modelling is achieved in two steps: (i) the state process modelling that enables to capture how states are generated, and (ii) the modelling of the autoregressive dynamics of the time series within each regime. In the latter step, a simple autoregressive framework such as the LAR model can be used. Generally, in step (i), the state process is modelled by a \textbf{discrete-valued Markov process}. In the current state-of-the-art literature, two categories of models can be distinguished.

In {\bf Hidden Regime-Switching Autoregressive (HRSAR) models}, the state process is hidden and is modelled by a Hidden Markov Process (HMP). This category of models has been introduced by \cite*{hamilton_1989_econometrica_nonstationary-timeseries-markov-switching-ar} in the context of United States's Gross National Product time series analysis. Several variants and extensions were subsequently designed.

In {\bf Observed Regime-Switching Autoregressive (ORSAR) models}, the state process is either observed or derived \textit{a priori}. In the latter case, a clustering algorithm is used before fitting the model, to extract the regimes. The clustering may either rely on endogenous variables ({\it i.e.}, the variables whose dynamics is observed through the time series) or on exogenous variables supposed to drive regime-switching. The recent work of \cite*{bessac_ailliot_cattiaux_et_al_2016_advanc-stat-climat-met-ocean_hidd-obs-regime-switch-ar-model} illustrates the application of these models to wind time series. 

When the state process is partially observed, which means that the system state is known at some random time-steps and unknown for the remaining time-steps, ORSAR models cannot be directly applied while HRSAR models are suboptimal in the sense that the observed states cannot be included.

To overcome these limitations, in this work, we propose a novel \textbf{regime-} \textbf{switching autoregressive} model that capitalizes on the observed states while the hidden states are inferred. We consider a special case of Markov process henceforth named \textbf{Markov Chain}. Our model is referred to as the Partially Hidden Markov Chain Linear AutoRegressive (PHMC-LAR) model. The PHMC-LAR model is a flexible parametric model that supplies a unification of HRSAR and ORSAR models when the state process is a Markov Chain. Thus, when the state process is fully observed, PHMC-LAR is reduced to ORSAR. Reversely, when the state process is fully hidden, PHMC-LAR instantiates as HRSAR. Beyond the unification aspect, we contribute to the machine learning literature through designing the underlying algorithmic machinery dedicated to effective and efficient PHMC-LAR model training. 

The main contributions of this paper are as follows:

\begin{enumerate}
	\item We propose a new regime-switching autoregressive model that integrates the states observed at some random time-steps. This model, referred to as PHMC-LAR, provides a unification of HRSAR and ORSAR models when the state process is modelled by a Markov Chain (MC).
	\item We propose a variant of the \textbf{Expectation-Maximization} (EM) \textbf{algorithm} that allows to learn the parameters of our model.
	\item Inference on hidden states is carried out by a variant of the \textbf{Viterbi algorithm}, adapted to take into account the observed states.
	\item Regarding the time series forecasting task, a prediction function is proposed. We distinguish between the case where the system state is known at forecast horizons from the case where it is latent. 
\end{enumerate}

The ability of our model to infer the hidden states and to make accurate predictions on time series, even when the observed states are unreliable, is investigated through experiments performed on synthetic data. Our work underlines the benefits of using partially observed states to decrease EM convergence times. This performance is obtained with no or practically no impact on the quality of hidden state inference, as from labelling percentages around 20$\%$-30$\%$; the prediction accuracy is also preserved above such percentage thresholds. For instance, for a training set of 100 sequences, with $70\%$ labelled states, the EM algorithm converges after 22 iterations on average against 62 on average for the unsupervised case. Moreover, performing fully supervised training with a proportion of ill-labelled states is also beneficial for EM convergence. For example, given a training set of size 100 annotated with a $70\%$-reliable labelling function, the EM algorithm converges after a single iteration against $67$ iterations for the unsupervised case. This offers promising prospects to enhance model selection for the PHMC-LAR model. Further experimentations also show the ability of our variant of the Viterbi algorithm to infer hidden states in partially-labelled sequences. In addition, while assessing the impact on predictions generated by incorporating labelled states in the training sequences, we also compared the situations where all states are unknown at forecast horizons to the situations where all states are known. Prediction errors are subdued at all horizons in the latter case (by 44\% on average), but contrasted horizons are still evidenced with low (respectively high) scores as in the former case. The constrast is kept constant whatever the percentage of observed states in the training set. Besides, we also point out the robustness of our model to labelling errors in inference task, when large training datasets and moderate labelling error rates are considered. Finally, the latter experiment highlights the remarkable robustness to error labelling in the prediction task, over the whole range of error rates.  

This paper is organized as follows. Related work is reviewed in Section \ref{related.work}. Section \ref{proposed.model} describes the PHMC-LAR model. Then a learning algorithm is derived in Section \ref{learning.algorithm}, to estimate the model parameters. Inference of the hidden states is addressed in Section \ref{inference}. Section \ref{forecasting} presents the time series forecasting procedure. Section \ref{experiments} depicts the experimental protocol that drove our experimentations on synthetic data, and discusses the results obtained. Section \ref{conclusion} concludes this paper.


\section{Related work}
\label{related.work}
This section first highlights the links between our proposal, PHMC-LAR, and the most closely related contributions of the literature. The PHMC-LAR combines a variant of the Hidden Markov Model (HMM), namely the Partially Hidden Markov Chain (PHMC), with the Linear AutoRegressive (LAR) model. The rest of this section reviews the two main models that compose the hybrid model proposed.

As mentioned in the introduction, the PHMC-LAR model unifies the HRSAR and ORSAR frameworks. However, the common thread between these latter frameworks is the implication of dependencies that drive the local dynamics within each regime. Therefore, the contributions of the literature most closely related to PHMC-LAR are also characterized by various local dynamics.

Several models closely related to HRSAR were proposed in the literature. The MS-AR model (Markov-switching AutoRegressive model) designed by \cite*{hamilton_1989_econometrica_nonstationary-timeseries-markov-switching-ar} combines ARIMA (AutoRegressive Integrated Moving Average) models with an HMM, to characterize changes in the parameters of an autoregressive process. The targeted application motivating the MS-AR model was economic analysis: the switch between fast growth and slow growth is governed by the outcome of the Markov process.

Further, \cite*{filardo_1994_j-of-busin-econo-stat_markov-switching-autoregress-time-varying-transit} incorporated time-varying transition probabilities between regimes in the MS-AR model. For instance, the resulting model was subsequently used to reproduce the cyclic patterns existing in climatic variables \citep{cardenas-gallo_sanchez-silva_akhavan-tabatabaei_et_al_2016_natural-hazards_markov-switching-autoregress-time-varying-transit_climate}. In parallel, the Hamilton's MS-AR model was also extended into a general dynamic linear model combined with Markov-switching \citep{kim_1994-j-of-economet-markov-switching-general-dynamic-lin_model}. Finally, Michalek and co-authors'work focused on a HRSAR model that integrates HMM with Moving Average (MA) models \citep{michalek_wagner_timmer_2000_ieee-trans-on-signal-process_arma-hmm-approx-estim}. In the same work, the parameter estimation approximation thus derived was generalized to deal with AutoRegressive Moving Average (ARMA) hybridized with HMM. Simulations of electrophysiological recordings showed that the derived estimators allow to recover the true dynamics where standard HMM fails. The model generalized by Michalek and collaborators, to integrate HMM with ARMA, was also applied to model human activity as time signals for activity early recognition \citep{li_fu_2012_conf-pattern-reco_arma-hmm-human-activity-recogn}.

More recently, a nonhomogeneous HRSAR model was developed to model wind time series \citep{ailliot_2015_jour-statist-planning-inferen_non-homog-hidden-markov-switching-models}. The aim was to acknowledge that the probability of switching from cyclonic conditions to anticyclonic conditions between time-steps $t$ and $t+1$ depends on the wind conditions at time-step $t$ at some given location off the French Brittany coast. A nonhomogeneous MS-AR (NHMS-AR) model was thus designed for this purpose.

To our knowledge, the investigations around ORSAR models are limited to the recent work of \cite*{bessac_ailliot_cattiaux_et_al_2016_advanc-stat-climat-met-ocean_hidd-obs-regime-switch-ar-model} which was applied to wind time series. Therein, observed regimes are derived by running a clustering procedure on the variables under study or on extra variables. Thus are identified the states, all distinct from one other, in which the data are homogeneous. Besides comparing the ORSAR models derived from various clustering procedures, Bessac and collaborators also compare the respective merits of HRSAR and ORSAR models on real-world and simulated data.

\subsection{Partially Hidden Markov Chain - PHMC($K$)}
\indent Hidden Markov models (HMMs) have been successfully used in such domains as natural language processing \citep{morwal_jahan_chopra_2012_journ-natural-lang-comput_HMM-nat-language-process}, handwriting recognition \citep{mouhcine_mustapha_zouhir_2018_journ-elsevier_handwritten-recognition-with-HMM}, speech emotion recognition \citep{schuller_rigoll_lang_2003_ic-multimed-expo_hmm-speech-emot-recogni}, human action recognition \citep{berg_reckordt_richter_2018_procedia-cirp_human-action-recogni-hmm} or renewable power prediction \citep{ghasvarian-jahromi_gharavian_mahdiani_2020_soft-comput_solar-power_predict-hmm}, to name but a few. 

HMM$(K$) is a flexible probabilistic \textit{framework} able to model complex hidden-regime-switching systems. It exactly possesses $K$ states where each state drives the specific behavior of an observed variable. This variable is itself modelled through a usual probability law such as a Gaussian law, for example. The system state process, which specifies the ongoing behavior of the latter observed variable at each time-step, is fully latent. Therefore, state inference is the main purpose of HMM models: the goal is to learn about the latent sequence of states from the observed behavior. This task is generally driven by Maximum A Posteriori (MAP) estimation implemented through the \textbf{Viterbi algorithm} \citep{forney_1973_journ-proceedings_viterbi-algo}. Importantly, the HMM framework satisfies the Markov property, which stipulates that the conditional probability distribution of the hidden state at time-step $t$, given the hidden states at previous time-steps $t' < t$, only depends on the hidden state at time-step $t-1$. Besides, the observed behavior at time-step $t$ solely depends on the hidden variable at time-step $t$. 

 
When dealing with systems in which the state process is partially observed or known, applying HMM would result in an important information loss in the sense that the observed states are ignored. To overcome this limitation, \cite*{scheffer-wrobel_2001_ecml-pkdd-workshop_partially-hidden-markov-models} have introduced the Partially Hidden Markov Chain (PHMC), which integrates partially observed states in the modelling process. The authors have proposed an active learning algorithm in which the user is asked to label difficult observations identified during model learning. More recently, \cite*{ramasso-denoeux_2013_jour-ieee-transact-fuzzy-sys_partially-hmms} have proposed a model that makes use of partial knowledge on HMM states. These authors have modelled the partial knowledge by a \textit{belief function} that specifies the probability of each state at each time-step. The works carried out by \cite*{ramasso-denoeux_2013_jour-ieee-transact-fuzzy-sys_partially-hmms} have shown that the use of partial knowledge on states accelerates HMM model learning.


\subsection{Linear AutoRegressive model - LAR($p$)}
An observed time series is considered to be one realization of a stochastic process. Time series analysis and forecasting thus require that the underlying stochastic process be modelled. The linear autoregressive (LAR) model is a stochastic model widely used for this purpose. A LAR model of order $p$ is a linear model in which the regressors are the $p$ past values of the variable, hence the term autoregression. Although the LAR model is conceptually simple and easy to learn, it can only be applied to \textbf{stationary time series}. When this condition is violated, \textit{model misspecification} issues arise. Nonetheless, it is well known that if the autoregressive coefficients of a LAR process are all less than one in module, then the process will be stationary. This is a necessary and sufficient condition which is tested through \textbf{unit root tests} \citep{phillips_perron_1988_biometrika_test-unit-root-time-series-regres, dickey_fuller_1979_journ-americ-stat-assoc_-ar-times-series-unit-root, kwiatkowski_phillips_schmidt_et_al_1992_journ-econometr_unit-root-test}.

In the LAR($p$) model, the hyper-parameter $p$ denotes the number of past observations to include in the prediction at time-step $t$. Two alternative methods are generally used to fix the value of $p$. The first one relies on a well-known property of the \textit{partial autocorrelation function} of the LAR($p$) model: the autocorrelation becomes null from lag $p+1$. The second method, more general, tests a range of candidate values for $p$, then selects the value that minimizes a model selection criterion such as the Bayesian information criterion (BIC) or the Akaike's information criterion (AIC).

\section{The PHMC-LAR model}
\label{proposed.model}

In this section, we explain how we have created a new regime-switching model called PHMC-LAR, based on the PHMC and LAR models. The section first introduces some notations. Then Subsection \ref{second.level} describes our proposal to model the state process by a PHMC model. Subsection \ref{first.level} details how, within each regime, the dynamics of the observed variable is governed by a LAR model. Thus, the bivariate process follows a PHMC-LAR model. 

\begin{figure}[t]
\centering
\begin{subfigure}[b]{0.38\textwidth} 
        \includegraphics[width=\textwidth, clip]{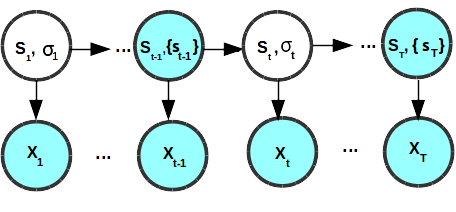}
 	\caption{}
 	\label{phmc.model}
 \end{subfigure}
 \hspace{5mm}
 \begin{subfigure}[b]{0.55\textwidth} 
        \includegraphics[width=\textwidth, clip]{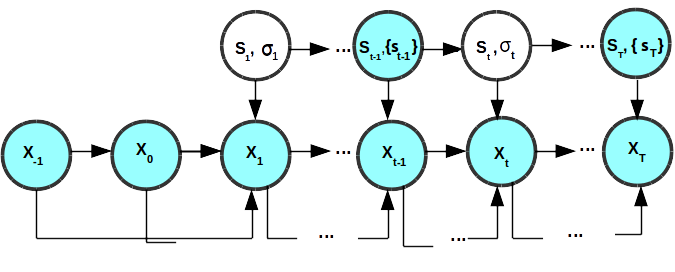}
 	\caption{}
 	\label{phmclar.conditional.ind.graph}
 \end{subfigure}
 \caption{The conditional independence graphs of the Partially Hidden Markov Chain and of the Partially Hidden Markov Chain Linear Autoregressive (PHMC-LAR) model, when the LAR order $p$ is equal to $2$. (a) PHMC model. (b) PHMC-LAR model. Observed states are shown in dark shade whereas hidden states are colored in light shade. When a state is observed, $\sigma_t$ is reduced to a singleton.}
 \label{phmc.phmclar.graphs}
 \end{figure}

To note, the fundamental difference between our model and the two other approaches identified in the same line \cite*{scheffer-wrobel_2001_ecml-pkdd-workshop_partially-hidden-markov-models,ramasso-denoeux_2013_jour-ieee-transact-fuzzy-sys_partially-hmms} is the autoregressive dynamics of our model (see Fig. \ref{phmc.phmclar.graphs}).

\subsection{Notations}
\hspace{4mm} $\bullet$ Symbol $:=$ stands for the {\it definition symbol}.
 
$\bullet$ $\bm{1}_A: \Omega \rightarrow \{0,1\}$ denotes the {\it indicator function} that indicates membership of an element in a subset $A$ of $\Omega$. As from now, $\bm{1}_A$ will be noted $\bm{1}_{\{ x \in A \}}$.

$\bullet$ $\{X_t\}_{t \in \mathbb{Z}}$ denotes a stochatic process. By convention, $X_{1-p}^0$ denotes the $p$ initial values of the time series $\{X_t\}$. For each $t \geq 1$, $X_{t-p}^{t-1}$ stands for the subseries $\{ X_{t-p}, X_{t-p+1} \cdots X_{t-1} \}$. 

$\bullet$ $\mathbf{x} = x_{1}^T$ represents an observed time series with $\mathbf{x}_0=x_{1-p}^0$ the corresponding initial values.  

$\bullet$ $\{ S_t \}_{t \in \mathbb{N}^*}$ denotes a state process depicting the temporal evolution of a regime-switching system where the set of states is $\mathbf{K} = \{1, 2, \dots, K\}$. In this paper, states are instantaneous, whereas a regime is a succession of identical states. We denote $\sigma_t$ the set of possible states at time-step $t$ with $\sigma_t = \mathbf{K}$ when $S_t$ is latent, and $\sigma_t = \{k\}$ when $S_t = k$, that is $k^{\text{th}}$ state is observed at time-step $t$.

$\bullet$ $\mathcal{M}_p(\mathbb{R})$ is the set of square matrices of order $p$ with real coefficients.  

$\bullet$ Symbols in bold represent nonscalar variables ({\it e.g.}, vectors).

\subsection{Modelling the state process}
\label{second.level}
Let $\{ (S_t, \sigma_t) \}$ the state process which is supposed to be partially observed. Remind that if $S_t=k$, {\it i.e.} $k^{\text{th}}$ state has been observed at time-step $t$, then $\sigma_t = \{k\}$; otherwise $\sigma_t = \mathbf{K}$, {\it i.e.} $S_t$ is latent. 

Let $\mathcal{R} = \{ k \in \mathbf{K} \,|\, \exists\, t \in \mathbb{N}^*, \sigma_t = \{k\} \}$, the set of states that have been observed at least once. We have $|\mathcal{R}| \le K$ where $K$ is the total number of states. Thus, $K - |\mathcal{R}|$ states are fully latent and depict the hidden dynamics of the system under study. It has to be underlined that it is difficult (it not sometimes impossible) to associate a physical interpretation to the hidden dynamics. Such an interpretation requires strong knowledge upon the studied system.

In the PHMC-LAR model, $\{(S_t, \sigma_t)\}$ is modelled by a $K$-state PHMC, parametrized by 
transition probabilities 

\noindent $a_{i,j} = P(S_t=j|S_{t-1}=i), \quad a_{i,j} \in [0, 1], \, \sum_{j=1}^K a_{i,j} = 1$

\noindent and stationary law $\pi_i = P(S_1 = i), \quad \pi_i \in [0, 1], \, \sum_{i=1}^K  \pi_i = 1$.

Let $\bm{\theta}^{(S)} = ( (\pi_i)_{i=1,...,K}, (a_{i,j})_{i,j=1,...,K})$ denote the set of parameters associated with the PHMC.

\subsection{Modelling the dynamics under each state}
\label{first.level} 
For each state $k \in \mathbf{K}$, $\{X_t\}$ is supposed to be \textbf{stationary} and modelled by a $p$-order LAR process defined as follows:

\begin{align}
\label{X.model}
X_t \,|\, X_{t-p}^{t-1}, S_t=k  \,&:=\, \phi_{0,k} + \sum_{i=1}^{p} \phi_{i,k} X_{t-i} + h_k \, \epsilon_t \quad \text{for} \quad t=1, \dots, T,
\end{align}{}

with $p$ the number of past values of $X_t$ to be used in modelling, $k$ the state at time-step $t$, $\bm{\mu}_k = (\phi_{0,k}, \, \phi_{1,k}, ..., \phi_{p,k})$ the intercept and autoregressive parameters associated with $k^{th}$ state, $h_k$ the standard deviation associated with $k^{th}$ state and $\{\epsilon_t\}$ the error terms.

It is important to underline that Eq. \ref{X.model} is not defined for the $p$ initial values denoted by $X_{1-p}^0$. These initial values are modelled by the initial law $g_0(x_{1-p}^0; \bm{\psi})$ parametrized by $\bm{\psi}$. For instance, $g_0$ can be a multivariate normal distribution $\mathcal{N}_p(\mathbf{m}, \mathbf{V})$ where $\mathbf{m} \in \mathbb{R}^p$ is the mean and $\mathbf{V} \in \mathcal{M}_p(\mathbb{R})$ is the variance-covariance matrix.

The $\epsilon_t$ terms are independent and identically distributed with zero mean and unit variance. Note that the law of $\{\epsilon_t\}$ and the conditional distribution $P(X_t|X_{t-p}^{t-1},$ $S_t=k; \, \bm{\mu}_k, h_k)$ belong to the same family. Usually, Gaussian white noises are used. In this case, the conditional distribution is Gaussian too, with mean and variance respectively equal to $\phi_{0,k} + \sum_{i=1}^p \phi_{i,k} X_{t-i}$ and $h_k^2$.

Let $\bm{\theta}^{(X,k)} = (\bm{\mu}_{k}, h_{k})$ the parameters of the LAR($p$) process associated with $k^{th}$ state. The law of $\{X_t\}$ is fully parametrized by $\bm{\theta}^{(X)} = (\bm{\theta}^{(X,k)})_{k=1, ...,K}$ and $\bm{\psi}$.

To note, as in \citep{scheffer-wrobel_2001_ecml-pkdd-workshop_partially-hidden-markov-models} and \citep{ramasso-denoeux_2013_jour-ieee-transact-fuzzy-sys_partially-hmms}, the PHMC-LAR model assumes that the same order $p$ is shared by all $| \mathbf{K} |$ LAR processes associated with the states in $\mathbf{K}$.

It has also to be highlighted that the state $S_t = k$ conditioning a LAR process of order $p$ on $X_t$ does not impose that the $p$ lagged values $X_{t-1}^{t-p}$ be observed under same state $k$. That is, the PHMC-LAR model may perfectly switch from regime to regime, and even from state to state, meanwhile keeping memory of values determined by previous regimes or states.

\section{Learning algorithm}
\label{learning.algorithm}
This section is dedicated to the presentation of an instance of the Expectation-Maximization (EM) algorithm, to estimate the PHMC-LAR parameters.
As seen in previous subsections, the PHMC component and the LAR components of our model are respectively parametrized by  $\bm{\theta}^{(S)}$ and $(\bm{\theta}^{(X)}, \bm{\psi})$. Then, the whole PHMC-LAR model is parametrized by $(\bm{\theta}, \bm{\psi})$ where $\bm{\theta} = (\bm{\theta}^{(S)}, \bm{\theta}^{(X)})$. Thus, PHMC-LAR learning consists in estimating $(\bm{\theta}, \bm{\psi})$ from a training dataset. 

Thanks to good statistical properties such as asymptotic efficiency, a maximum likelihood estimator (MLE) is considered. However, for models with hidden variables like ours, MLE computation results in an untractable problem. To address this issue, the Expectation-Maximization (EM) algorithm is generally used, in order to approximate a set of parameters that locally maximizes the likelihood function. EM was introduced by \cite*{baum-petrie-soules_1970_jour-annals-math_em-for-hmc-models} to cope with Hidden Markov Model learning. This version was further extended by \cite{dempster-laird-rubin_1977_jour-royal-stat-society_EM-for-incomplete-data} into the versatile EM algorithm, to handle parameter estimation in a more general framework. EM has also been applied to autoregressive Markov-switching models \citep{hamilton_1990_journ-econometrics_time_series-regime-changes} and PHMC models \citep{scheffer-wrobel_2001_ecml-pkdd-workshop_partially-hidden-markov-models, ramasso-denoeux_2013_jour-ieee-transact-fuzzy-sys_partially-hmms}.

We propose to learn the PHMC-LAR model through a dedicated instance of the EM algorithm. To fix ideas, in subsection \ref{particular.case}, we first consider the case where the model is trained in a univariate context, that is considering a unique couple of data ($x_{t=1-p}^T, \, \sigma_{t=1}^T$), with $x_{t=1-p}^T$ a realization of $\{ X_t \}$ and $\sigma_t$ the set of possible states at time-step $t$. Then, the general multivariate case of $N$ independent couples of data $(\mathbf{x}^{(1)},\Sigma^{(1)})$, $\dots$, $(\mathbf{x}^{(N)},\Sigma^{(N)})$ will be presented in subsection \ref{general.case}.

\subsection{Particular case: univariate scheme}
\label{particular.case}

Let $\mathbf{x} = x_{1-p}^T$ the observed time series with $x_{1-p}^0$ the initial values of the autoregressive process. Let $\Sigma = \sigma_{t=1}^T$, further simplified into $\sigma_{1}^T$, where $\sigma_t$ stands for the set of possible states at time-step $t$. Let $(S_1^T, \Sigma)$ the state process (partially observed) of $\mathbf{x}$ with $\sigma_t = \mathbf{K}$ if $S_t$ is hidden, and $\sigma_t = \{k\}$ if state $k$ is observed at time-step $t$.

MLE is implemented by maximizing the expectation (with respect to the latent variables) of the \textbf{complete data likelihood}.
Complete data likelihood is further referred to as $\mathcal{L}^c$. $\mathcal{L}^c$ denotes the evidence/likelihood of the training data when latent/hidden variables are supposed to be known. $\mathcal{L}^c$ writes as follows:

\begin{align}
    \mathcal{L}^c (\bm{\theta}, \bm{\psi}) &= P(X_{1-p}^T=x_{1-p}^T, S_1^T=s_1^T; \, \bm{\theta}, \bm{\psi} ) \nonumber \\
    &= P(X_1^T=x_1^T, S_1^T=s_1^T \,|\, X_{1-p}^0=x_{1-p}^0; \, \bm{\theta}) \times P(X_{1-p}^0=x_{1-p}^0; \, \bm{\psi}) \nonumber  \\
\label{complete.data.like}
    &= \mathcal{L}^c_c (\bm{\theta}) \times g_0(x_{1-p}^0; \, \bm{\psi}),
\end{align}

with $\mathcal{L}_c^c$ the \textbf{conditional complete data likelihood} and $g_0$ the initial law of $X_t$.

When the expectation of $\mathcal{L}^c$ with respect to the partially hidden states is calculated, term $g_0(x_{1-p}^0; \, \bm{\psi})$ in Eq. \ref{complete.data.like} can be taken out of the expectation since it does not depend on the states:

{\small
\begin{equation}
\label{expect.complete.data.like}
\begin{split}
	\mathbb{E}_{P(S_1^T \,|\, X_{1-p}^{T}=x_{1-p}^{T}, \Sigma; \, \bm{\theta})} [\mathcal{L}^c(\bm{\theta}, \bm{\psi})] &= \mathbb{E}_{P(S_1^T \,|\, X_{1-p}^{T}=x_{1-p}^{T}, \Sigma; \, \bm{\theta})} [\mathcal{L}^c_c (\bm{\theta})] \times g_0(x_{1-p}^0; \, \bm{\psi}),
\end{split}
\end{equation}
} 

where $P(S_1^T \,| \,X_{1-p}^{T}=x_{1-p}^{T}, \Sigma; \, \bm{\bm{\theta}})$ is the \textit{posterior probability} of partially hidden states $(S_1^T, \Sigma)$.\\

Then, by considering the logarithmic scale, Eq. \ref{expect.complete.data.like} can be separately maximized with respect to $\bm{\theta}$ and $\bm{\psi}$:
 
\begin{align}
\label{psi.max}
	\hat{\bm{\psi}} &= \underset{\bm{\psi}}{\arg\max} \, \ln \left( g_0(x_{1-p}^0; \, \bm{\psi}) \right), \\
\label{theta.max}
	\hat{\bm{\theta}} &= \underset{\bm{\theta}}{\arg\max} \, \ln \left( \mathbb{E}_{P(S_1^T \,|\, X_{1-p}^{T}=x_{1-p}^{T}, \Sigma; \, \bm{\theta})} [\mathcal{L}_c^c(\bm{\theta})] \right).
\end{align}

It has to be noted that Eq. \ref{psi.max} is a simple probability observation problem. In contrast, because of the hidden states, maximization with respect to $\bm{\theta}$ (Eq. \ref{theta.max}) is carried out by an instance of the EM algorithm. 

\indent EM is an iterative algorithm that alternates between E(xpectation) step and M(aximization) step. At iteration $n$, we obtain:

\begin{align}
	\label{e.step}
     \text{E-step} & \quad Q(\bm{\theta}, \hat{\bm{\theta}}_{n-1}) = \mathbb{E}_{P(S_1^T \,|\, X_{1-p}^{T}=x_{1-p}^{T}, \Sigma; \, \hat{\bm{\theta}}_{n-1})} [\ln \mathcal{L}_c^c (\bm{\theta})], \\
     \label{m.step}
     \text{M-step} & \quad \hat{\bm{\theta}}_n = \underset{\bm{\theta}}{\arg\max} \,\ Q(\bm{\theta}, \hat{\bm{\theta}}_{n-1}),
\end{align}

with $P(S_1^T \,| \,X_{1-p}^{T}=x_{1-p}^{T}, \Sigma; \, \hat{\bm{\bm{\theta}}}_{n-1})$ the \textit{posterior probability} of partially hidden states $(S_1^T, \Sigma)$ at iteration $n-1$. 

The rest of this Subsection details the two EM steps.

\subsubsection{Step E of EM}
In this step, the quantity $Q(\bm{\theta}, \hat{\bm{\theta}}_{n-1})$ (Eq. \ref{e.step}) is computed. Following the conditional independence graph of the PHMC-LAR model (see Fig. \ref{phmclar.conditional.ind.graph}), the conditional complete data likelihood writes:

\begin{equation}
\label{cond.comp.data.like}
\begin{split}
\mathcal{L}_c^c (\bm{\bm{\theta}}) 
    &= P(X_1^T=x_1^T, S_1^T=s_1^T \,|\, X_{1-p}^0; \, \bm{\theta})  \\
    &= P(S_1=s_1; \, \bm{\theta}^{(S)}) \prod_{t=2}^T P(S_t=s_t|S_{t-1}=s_{t-1}; \, \bm{\theta}^{(S)})\\
    & \quad \prod_{t=1}^T P(X_t=x_t|X_{t-p}^{t-1}=x_{t-p}^{t-1}, S_t =s_t; \, \bm{\theta}^{(X,s_t)}),
\end{split}
\end{equation}

with $\bm{\theta}^{(X,k)}$ the parameters of the LAR process associated with $k^{th}$ state and $P(X_t=x_t \, | \, X_{t-p}^{t-1}, S_t =k; \, \bm{\theta}^{(X,k)})$ the conditional law of $X_t$ within $k$. \\

Notice that the terms in Eq. \ref{cond.comp.data.like} depend on either a single state $S_t$ or two consecutive states $S_t, S_{t-1}$. In this same equation, products are replaced by sums when considering the logarithm scale. Then $\ln \mathcal{L}_c^c (\bm{\bm{\theta}})$ is substituted in Eq. \ref{e.step} and the expectation with respect to the posterior probability of state process is developed. After some integrations, we find that $Q(\bm{\theta}, \hat{\bm{\theta}}_{n-1})$ only depends on the following probabilities:

\begin{equation}
\label{xi.definition}
\begin{split}
  &\xi_t(k, \ell) = P(S_{t-1}=k, S_{t}=\ell \, | \, X_{1-p}^T = x_{1-p}^T, \Sigma; \, \hat{\bm{\theta}}_{n-1}),\\
  &\text{for} \quad t=2, \dots, T, \quad 1 \le k, \,\ell \le K.\\ 
\end{split}
\end{equation}

\begin{equation}
\label{gamma.defintion}
\begin{split}
  &\gamma_t(\ell) = P(S_{t}=\ell \, | \, X_{1-p}^T = x_{1-p}^T, \Sigma; \, \hat{\bm{\theta}}_{n-1}),\\
  &\text{for} \quad t=2, \dots, T, \quad 1 \le \ell \le K.\\
\end{split}
\end{equation}

Therefore, the E-step is reduced to computing these probabilities. To this end, we have derived a \textit{backward-forward-backward} procedure as an extension of the forward-backward algorithm, one of the ingredients of the Baum-Welsh algorithm \citep{dempster-laird-rubin_1977_jour-royal-stat-society_EM-for-incomplete-data}. The backward-forward-backward algorithm was initially proposed by \cite*{scheffer-wrobel_2001_ecml-pkdd-workshop_partially-hidden-markov-models} for the purpose of PHMC model learning. We have adapted this algorithm to PHMC-LAR models by taking into consideration the autoregressive dynamics. The details about the adapted \textit{backward-forward-backward} algorithm are given in Appendix \ref{BFB}.

\subsubsection{Step M of EM}
At iteration $n$, this step consists in maximizing $Q(\bm{\theta}, \hat{\bm{\theta}}_{n-1})$ with respect to parameters $\bm{\theta} = (\bm{\theta}^{(S)}, \bm{\theta}^{(X)})$. It is straightforward to show that $Q(\bm{\theta}, \hat{\bm{\theta}}_{n-1})$ can be decomposed as follows:
$$Q(\bm{\theta}, \hat{\bm{\theta}}_{n-1}) = Q_S(\bm{\theta}^{(S)}, \hat{\bm{\theta}}_{n-1}) +  Q_X(\bm{\theta}^{(X)}, \hat{\bm{\theta}}_{n-1}),$$
where $Q_S$ (respectively $Q_X$) only depends on parameters $\bm{\theta}_S$ (respectively $\bm{\theta}_X$). Therefore, $Q_S$ and $Q_X$ can be maximized apart:

\begin{equation}
\label{theta.S}
    \hat{\bm{\theta}}_n^{(S)} = \underset{\bm{\theta}^{(S)}}{\arg\max} \,\ Q_S(\bm{\theta}^{(S)}, \hat{\bm{\theta}}_{n-1}).
\end{equation}

\begin{equation}
\label{theta.X}
    \hat{\bm{\theta}}_n^{(X)} = \underset{\bm{\theta}^{(X)}}{\arg\max} \,\ Q_X(\bm{\theta}^{(X)}, \hat{\bm{\theta}}_{n-1}).
\end{equation}

The analytical expressions of $Q_S$ and $Q_X$ are given in Appendix \ref{decomposition.of.Q}. Cancelling the first derivative of $Q_S(\bm{\theta}^{(S)}, \hat{\bm{\theta}}_{n-1})$ provides the analytical expression of $\hat{\bm{\theta}}_n^{(S)}$. In contrast, it is not possible to derive the analytical expression for $\hat{\bm{\theta}}_n^{(X)}$. That is why $Q_X(\bm{\theta}^{(X)}, \hat{\bm{\theta}}_{n-1})$ has to be maximized relying on a numerical optimization method ({\it e.g.}, the quasi-Newton method).

\subsection{General case: multivariate scheme}
\label{general.case}
We now consider the general case in which PHMC-LAR model is learnt from $N$ independent couples of data $(\mathbf{x}^{(1)},\Sigma^{(1)})$, $\dots$, $(\mathbf{x}^{(N)},\Sigma^{(N)})$, with $\mathbf{x}_0^{(1)}, \dots, \mathbf{x}_0^{(N)}$ the associated initial values and $(\mathbf{S}^{(1)}, \Sigma^{(1)}), \dots, (\mathbf{S}^{(N)}, \Sigma^{(N)})$ the corresponding state processes. It has to be noted that time series $\mathbf{x}^{(i)}$'s can have different lengths while their respective initial vectors have a common size ($\mathbf{x}_0^{(i)} \in \mathbb{R}^p$, with $p$ the autoregressive order).

In this case, the MLE estimates of parameters ($\bm{\theta}, \bm{\psi}$) are defined as follows:

\begin{align}
\label{g.0.param.learning} 
    \hat{\bm{\psi}} &= \underset{\bm{\psi}}{\arg\max} \, \sum_{i=1}^N \ln \left( g_0(\mathbf{x}_0^{(i)}; \, \bm{\psi}) \right), \\
\label{theta.learning}
    \hat{\bm{\theta}} &= \underset{\bm{\theta}}{\arg\max} \, \ln \left( \mathbb{E}_{P_S} [\mathcal{L}_c^c(\bm{\theta})] \right),
\end{align}


with $P_S = P(\mathbf{S}^{(1)}, \dots, \mathbf{S}^{(N)} \,|\, \mathbf{X}^{(1)}=\mathbf{x}^{(1)},  \dots, \mathbf{X}^{(N)}=\mathbf{x}^{(N)}, \mathbf{X}^{(1)}_0=\mathbf{x}^{(1)}_0, \dots,$ $\mathbf{X}^{(N)}_0=\mathbf{x}^{(N)}_0, \Sigma^{(1)}, \dots, \Sigma^{(N)}; \, \bm{\theta})$ the \textit{posterior distribution} of partially hidden states $(\mathbf{S}^{(1)}, \Sigma^{(1)}), \dots, (\mathbf{S}^{(N)}, \Sigma^{(N)})$.

Thus, the conditional complete data likelihood $\mathcal{L}_c^c (\bm{\bm{\theta}}) $ defined in Eq. \ref{cond.comp.data.like} becomes:


{\small 
\begin{equation}
\begin{split}
\label{log.likelihood.several.seq}
\mathcal{L}_c^c (\bm{\bm{\theta}}) 
    &= \prod_{i=1}^N P \left(\mathbf{X}^{(i)}=\mathbf{x}^{(i)}, \mathbf{S}^{(i)}=\mathbf{s}^{(i)} \,|\, \mathbf{X}_0^{(i)}; \, \bm{\theta} \right)  \\
    &= \prod_{i=1}^N  \left[ P \left(S^{(i)}_1=s^{(i)}_1; \, \bm{\theta}^{(S)} \right) \prod_{t=2}^{T_i} P \left(S_t^{(i)}=s_t^{(i)} \,|\, S_{t-1}^{(i)}=s_{t-1}^{(i)}; \, \bm{\theta}^{(S)} \right) \, \right.\\
    & \qquad \quad \times \left. \prod_{t=1}^{T_i} P \left(X_t^{(i)}=x_t^{(i)} \,|\, [X^{(i)}]_{t-p}^{t-1}=[x^{(i)}]_{t-p}^{t-1}, \, S_t^{(i)} =s_t^{(i)}; \, \bm{\theta}^{(X,s_t^{(i)})} \right) \right].
\end{split}
\end{equation}
} 
 
As in Eq. \ref{psi.max}, Eq. \ref{g.0.param.learning} is a simple probability observation problem. When $g_0$ is a multivariate normal distribution $\mathcal{N}_p(\mathbf{m})$, with mean $\mathbf{m} \in \mathbb{R}^p$, variance-covariance matrix $\mathbf{V} \in \mathcal{M}_p(\mathbb{R})$ and $\bm{\psi} = (\mathbf{m}, \mathbf{V})$, we can show that

\begin{equation}
\hat{\mathbf{m}} = \frac{1}{N} \sum_{i=1}^N \mathbf{x}_0^{(i)}, \quad
    \hat{\mathbf{V}} = \frac{1}{N} \sum_{i=1}^N \, (\mathbf{x}_0^{(i)} - \hat{\mathbf{m}}) \, (\mathbf{x}_0^{(i)} - \hat{\mathbf{m}})^{'},
\end{equation}

where ${'}$ stands for matrix transposition.

Equation \ref{theta.learning} is maximized using the instance of EM presented in Subsection \ref{particular.case}. At each iteration $n$, the E-step consists in computing probabilities $\xi_t^{(i)}$ (Eq. \ref{xi.definition}) by running the \textit{backward-forward-backward} algorithm presented in Appendix \ref{BFB}, on data $(\mathbf{x}^{(i)}, \Sigma^{(i)})$. Thus, the expectation $Q(\bm{\theta}, \hat{\bm{\theta}}_{n-1})$ (Eq. \ref{e.step}) can be computed.

Then, in the M-step, $Q(\bm{\theta}, \hat{\bm{\theta}}_{n-1})$ is maximized with respect to $\bm{\theta} = (\bm{\theta}^{(S)}, \bm{\theta}^{(X)})$ following Eq. \ref{theta.S}-\ref{theta.X}. We obtain the following formula for $\hat{\bm{\theta}}^{(S)}_n$:

\begin{equation}
 \label{phmc.mle.1}
   \hat{a}_{k,l}^{(n)} = \frac{ \sum_{i=1}^N \sum_{t=2}^{T_i} \xi_t^{(i)}(k,l) } { \sum_{i=1}^N \sum_{t=1}^{T_i} \gamma_t^{(i)} (k)}, \quad \text{for} \quad 1 \le k,l \le K.
\end{equation}

\begin{equation}
\label{phmc.mle.2}
    \hat{\pi}_l^{(n)}    = \frac{ \sum_{i=1}^N \gamma_1^{(i)}(l) }{N}, \quad \text{for} \quad 1 \le l \le K,
\end{equation}

\begin{equation}
\label{gamma_general}
\begin{split}
    &\text{with} \quad \gamma_t^{(i)}(s) = P(S_t^{(i)}=s \,|\, [X^{(i)}]_{1-p}^{T_i} = x_{1-p}^{T_i}, \Sigma^{(i)}; \, \hat{\bm{\theta}}_{n-1}) = \sum_{j=1}^K \xi_t^{(i)}(j,s),\\
    &\text{for} \quad t=2, \dots, T_i,
\end{split}
\end{equation}
\begin{equation}
\label{gamma_one}
\begin{split}
    \text{and} \quad \gamma_1^{(i)}(s) &= P(S_1^{(i)}=s \,|\, [X^{(i)}]_{1-p}^{T_i} = x_{1-p}^{T_i}, \Sigma^{(i)}; \, \hat{\bm{\theta}}_{n-1}) = \sum_{j=1}^K \xi_2^{(i)}(s,j).
\end{split}
\end{equation}

Note that $\hat{\bm{\theta}}^{(X)}_n$ is computed through numerical optimization, for instance by using the quasi-Newton method.

Algorithm \ref{em.algorithm} sums up the instance of EM proposed for PHMC-LAR parameter learning.

\vspace{0.3cm}
\begin{algorithm}[H]
\begin{algorithmic}[1]
 \STATE \textbf{Input}: \textit{data} $(\mathbf{x}^{(1)}, \Sigma^{(1)}), \dots, (\mathbf{x}^{(N)}, \Sigma^{(N)})$, precision $\kappa$, maximal number of iterations $max_{iter}$ 
 \STATE Initialization: $\hat{\bm{\theta}}^{(0)}$ randomly chosen
 \STATE $n \leftarrow 1$ 
 \REPEAT
 	\STATE E-step
 	\STATE \hspace{0.4cm} For each couple $(\mathbf{x}^{(i)}, \Sigma^{(i)})$, $i= 1, \dots, N$
 	\STATE \hspace{0.4cm} Compute $\xi_t^{(i)}$ by running the \textit{backward-forward-backward} algorithm on $(\mathbf{x}^{(i)}, \Sigma^{(i)})$ 
    \STATE M-step
    		\STATE \hspace{0.4cm} M-S : compute $\hat{\bm{\theta}}_n^{(S)}$ from Eq. \ref{phmc.mle.1}-\ref{phmc.mle.2}
 		\STATE \hspace{0.4cm} M-X : compute $\hat{\bm{\theta}}_n^{(X)}$ by numerical optimization of $Q_X(\bm{\theta}^{(X)}, \hat{\bm{\theta}}_{n-1})$
        \STATE $incr(n)$
 \UNTIL{($|\hat{\bm{\theta}}^{(n)} - \hat{\bm{\theta}}^{(n-1)}| < \kappa$) or ($n> max_{iter}$)} 
        \STATE /* parameters stay roughly stable between two successive iterations, */
        \STATE /* or the maximum number of iterations is reached \verb^               ^*/                                             
\end{algorithmic}
\caption{EM algorithm for PHMC-LAR model training}
\label{em.algorithm}
\end{algorithm}

\vspace{0.3cm}
It is well known that the EM algorithm is sensitive to the choice of the starting point $\hat{\bm{\theta}}^{(0)}$ as regards the risk of attraction in a local maximum. In practice, several initial values are tested and the model that provides the highest likelihood is chosen. In this work, the initialization procedure presented in Algorithm \ref{em.init} is used.

\vspace{0.3cm}
\begin{algorithm}[H]
\begin{algorithmic}[1]
\STATE \textbf{Input}: $L$, precision $\kappa$, maximum number of iterations $max_{iter}$
\STATE Let $\hat{\bm{\theta}}^{(0,1)}, \dots, \hat{\bm{\theta}}^{(0, L)}$ initial values randomly chosen.
\STATE For each $\hat{\bm{\theta}}^{(0,j)}$, EM is run with parameters $\kappa$ and $max_{iter}$.
\STATE Then, $\hat{\bm{\theta}}^{(0)}$ is fixed as the estimated parameters that provide the highest likelihood across the $L$ restarts.
\end{algorithmic}
\caption{EM initialization for PHMC-LAR model training}
\label{em.init}
\end{algorithm}

\section{Hidden state inference}
\label{inference}
In HMM modelling, after a model is learnt, inference consists in finding the state sequence that maximizes the likelihood of a given observed sequence. This is equivalent to solve a \textit{Maximum A Posteriori} (MAP) problem. The Greedy search method that enumerates all combinations of states requires $\mathcal{O}(K^T)$ operations, where $K$ is the number of states and $T$ is the sequence length. The \textbf{Viterbi algorithm} designed by \cite*{forney_1973_journ-proceedings_viterbi-algo} computes the optimal state sequence in $\mathcal{O}(TK^2)$ operations.

In this section, we propose a variant of the Viterbi algorithm that takes into account the observed states of the PHMC-LAR model. Thus, the hidden states are inferred given the observed states and the given observation sequence.

Let $\hat{\bm{\theta}}$ the MLE parameter estimates of the PHMC-LAR model trained on a given dataset. Let $\mathbf{x} = x_{1}^T$ an observed time series and $\mathbf{x}_0=x_{1-p}^0$ the corresponding initial values. Let $\Sigma = \sigma_{t=1}^T$ the possible states at each time-step with $\sigma_t = \{k\}$ if $k^{th}$ regime is observed at time-step $t$, and $\sigma_t = \mathbf{K}$ if the state process is latent at that time-step. Let $(\mathbf{S}, \Sigma)$ the partially hidden state process associated with this time series.

We search the optimal state sequence $\mathbf{z}^* = (z_1^*, \dots, z_T^*)$ that maximizes the posterior probability $P(\mathbf{S}=\mathbf{z} \,|\, \mathbf{X}=\mathbf{x}, \mathbf{X}_0=\mathbf{x}_0, \Sigma; \, \hat{\bm{\theta}})$. Thanks to Bayes' rule, maximizing this posterior probability is equivalent to maximizing the joint probability $P(\mathbf{S}=\mathbf{z}, \mathbf{X}=\mathbf{x} \,|\, \mathbf{X}_0=\mathbf{x}_0, \Sigma; \, \hat{\bm{\theta}})$:

\begin{equation}
\label{joint.prob}
P(\mathbf{S}=\mathbf{z} \,|\, \mathbf{X}=\mathbf{x}, \, \mathbf{X}_0=\mathbf{x}_0 ; \, \hat{\bm{\theta}}^{(X)}) = \frac{P(\mathbf{S}=\mathbf{z}, \mathbf{X}=\mathbf{x} \,|\, \mathbf{X}_0=\mathbf{x}_0 \, \Sigma; \, \hat{\bm{\theta}})}{P(\mathbf{X}=\mathbf{x} \,|\,\mathbf{X}_0=\mathbf{x}_0, \Sigma; \, \hat{\bm{\theta}}^{(S)})}.
\end{equation}

\begin{equation}
\label{z.opt}
\mathbf{z}^* = \underset{\mathbf{z} \, \in \, \mathbf{K}^T}{\arg\max} \, P(\mathbf{S}=\mathbf{z}, \mathbf{X}=\mathbf{x} \,|\, \mathbf{X}_0=\mathbf{x}_0, \Sigma; \, \hat{\bm{\theta}}),
\end{equation}

where $\mathbf{K} = \{1, 2, \dots, K\}$ is the set of possible states.

Note that the probability of a given state sequence is null if there is at least a time-step $t$ such that $z_t \notin \sigma_t$, that is if state $z_t$ is not allowed at time-step $t$. A consequence is that $\mathbf{z}^*$ must coincide with the observed states if there are any.

Following the dynamic programming paradigm, the Viterbi algorithm makes it possible to retrieve $\mathbf{z}^*$ by splitting the initial problem into subproblems and solving this set of smaller problems. Let $\delta_t(\ell; \, \hat{\bm{\theta}})$ the maximal probability of subsequence $(z_1, \dots, z_t = \ell)$ that ends within regime $\ell$:

\begin{equation}
\label{viterbi.delta}
\begin{split}
 &\delta_t(\ell; \, \hat{\bm{\theta}}) = \underset{z_1, \dots, z_{t-1} \, \in \mathbf{K}^{t-1}}{\max} \, P(X_1^t=x_1^t, \, S_1^{t-1}=z_1^{t-1}, \, S_t=\ell \, | \, \mathbf{X}_0= \mathbf{x}_0, \, \sigma_1^t; \, \hat{\bm{\theta}}),\\
 &\text{for} \quad t = 1, 2, \dots T.
\end{split}
\end{equation}

\indent These probabilities are iteratively computed as follows:

At first time-step,
\begin{equation}
\label{delta.1}
\delta_1(\ell; \, \hat{\bm{\theta}}) = P(X_1=x_1 \,|\, \mathbf{X}_0= \mathbf{x}_0, \, S_1=\ell;  \, \bm{\theta}^{(X,\ell)}) \times P(S_1=\ell \,|\, \sigma_1; \, \hat{\bm{\theta}}^{(S)})
\end{equation}

where
\begin{equation*}
\label{delta.1}
\begin{split}
P(S_1=\ell \,|\, \sigma_1; \, \hat{\bm{\theta}}^{(S)}) &= \left\{ 
        \begin{aligned}
            \hat{\pi}_l \quad &\text{ if } \ell \in \sigma_1 \\
            0           \quad &\text{ otherwise.}
        \end{aligned}
    \right.
\end{split}{}
\end{equation*}

For $t=2, \dots, T$ we have
\begin{equation}
\label{delta.t}
 \begin{split}
 \delta_t(\ell; \, \hat{\bm{\theta}}) &= \underset{k}{\max} \, \left[\delta_{t-1}(k; \, \hat{\bm{\theta}}) \, P(S_t=\ell \,|\, S_{t-1}=k, \, \sigma_t; \, \hat{\bm{\theta}}^{(S)}) \right]\\
    &\quad \times  \, P(X_t=x_t \,|\, X_1^{t-1}=x_1^{t-1}, \mathbf{X}_0 = \mathbf{x}_0, \, S_t=\ell; \, \bm{\theta}^{(X,\ell)}),
 \end{split}
\end{equation}

with 
\begin{equation*}
\label{}
\begin{split}
    P(S_t=\ell \,|\, S_{t-1}=k, \sigma_t; \hat{\bm{\theta}}^{(S)}) &= \left\{ 
        \begin{aligned}
            \hat{a}_{k,\ell}   \quad &\text{ if } \ell \in \sigma_t \text{ and } k \in \sigma_{t-1} \\
                0          \quad &\text{ otherwise.}
        \end{aligned}
    \right.
\end{split}
\end{equation*}{}

Since the maximal probability of the complete state sequence, that is the maximum for the probability expressed in Eq. \ref{joint.prob}, also writes:

\begin{align}
    P^* = \underset{\ell}{\max} \, \delta_T(\ell; \, \hat{\bm{\theta}}),
\end{align}

the optimal sequence $\mathbf{z}^*$, defined in Eq. \ref{z.opt} is retrieved by backtracking as follows:

\begin{align}
    z_t^* &= \underset{\ell}{\arg\max}
    \left\{ 
        \begin{aligned}
            \delta_T(\ell; \, \hat{\bm{\theta}})  \quad \quad \quad  &\text{ for } t = T \\
            \delta_t(\ell; \, \hat{\bm{\theta}}) \times \hat{a}_{\ell,z_{t+1}^*} \quad &\text{ for } t = T-1, \dots, 1.
        \end{aligned}
    \right.
\end{align} 

\section{Forecasting}
\label{forecasting}
Forecasting for a time series consists in predicting future values based on past values. Let us consider a PHMC-LAR model trained on a sequence observed up to time-step $T$, and $\bm {\hat{\theta}}$ the corresponding parameters. Let $\sigma_{T+1}, \dots, \sigma_{T+h}$ the set of possible states from time-step $T+1$ to time-step $T+h$.

\indent The optimal prediction of $X_{T+h}$ (with respect to mean squared error) is the conditional mean $\mathbb{E}[X_{T+h} \, | \, X_{1-p}^{T}=x_{1-p}^{T}, \, \sigma_{T+1}^{T+h}; \, \bm {\hat{\theta}}]$, which writes as follows:

\begin{equation}
\label{forecasting.function}
\begin{split}
    \hat{X}_{T+h} 
    &= \sum_{k=1}^K P(S_{T+h} = k \, | \, X_{1-p}^{T}=x_{1-p}^{T}, \, \sigma_{T+1}^{T+h}; \, \bm {\hat{\theta}})\\
    & \qquad \quad \mathbb{E}[X_{T+h} \,|\, X_{T+h-p}^{T+h-1}=x_{T+h-p}^{T+h-1}, S_{T+h}=k; \, \bm {\hat{\theta}}]\\
    &= \sum_{k=1}^K P(S_{T+h} = k \, | \, X_{1-p}^{T}=x_{1-p}^{T}, \, \sigma_{T+1}^{T+h}; \, \bm {\hat{\theta}}) \, \left(\mathbf{y}_{T+h} \, \hat{\bm{\mu}}_k^{'} \right),
\end{split}
\end{equation}

with $\mathbf{y}_{T+h} = (1, x_{T+h-1}, \dots, x_{T+h-p})$, $\hat{\bm{\mu}}_k = (\phi_{0,k}, \phi_{1,k}, \dots, \phi_{p,k})$ the intercept and  autoregressive parameters associated with $k^{th}$ state, and $'$ denoting matrix transposition.

Equation \ref{forecasting.function} depends on \textbf{smoothed probabilities} $\bar{\gamma}(i, s) = P(S_{T+i}=s \,|\, X_{1-p}^{T}=x_{1-p}^{T}, \, \sigma_{T+1}^{T+i}; \, \bm{\hat{\theta}})$, which are recursively computed as follows:

\begin{equation}
\label{prediction.probs}
\left\{ 
  \begin{aligned}
	 \bar{\gamma}(0, s) &= P(S_{T}=s \,|\, X_{1-p}^{T}=x_{1-p}^{T}; \, \bm{\hat{\theta}})  = \gamma_T(s), \\
     \bar{\gamma}(i, s)  &=    \sum_{\ell=1}^K \, \hat{a}_{\ell,s} \, \bar{\gamma}(i-1, \, \ell) \quad \text{ if } \sigma_{T+i} = \mathbf{K}, \\
     \bar{\gamma}(i, s)  &= 1  \qquad \qquad \qquad \qquad \text{ if } \sigma_{T+i} = \{\ell\} \text{ and } s = \ell, \\
     \bar{\gamma}(i, s)  &= 0  \qquad \qquad \qquad \qquad \text{ if } \sigma_{T+i} = \{ \ell\} \text{ and } s \neq \ell,  
  \end{aligned}
\right.
\end{equation}

for $i=1,  \dots, h$, $s \in \mathbf{K}$ and $\gamma_{T}(l)$ defined in Eq. \ref{gamma_general}.

From Eq. \ref{forecasting.function} and \ref{prediction.probs}, we can notice that if state $s$ is observed at time-step $T+h$ ({\it i.e.} $\sigma_{T+h} = \{s\}$), then prediction $\hat{X}_{T+h}$ equals the conditional mean of the LAR process associated with this state (since $\bar{\gamma}(h, k) = 0$ for $k \notin \sigma_{T+h}$). In contrast, if state process is latent at time-step $T+h$ ({\it i.e.,} $\sigma_{T+h} = \mathbf{K}$), $\hat{X}_{T+h}$ is computed as the weighted sum of the conditional means of all states, with probabilities $\bar{\gamma}(h, k)$ as weights.

Note that for $h=1$, the past values of the time series required in Eq. \ref{forecasting.function} are known. In contrast, for $h > 1$, the intermediate predictions $\hat{X}_{T+1}, \dots, \hat{X}_{T+h-1}$ are used in order to feed the autoregressive dynamics of the PHMC-LAR framework.


\section{Experiments}
\label{experiments}
The aim of this section is two-fold: (i) assess the ability of PHMC-LAR model to infer the hidden states, (ii) evaluate prediction accuracy. These evaluations were achieved on simulated data, following two experimental settings. On the one hand, we varied the percentage of observed states in training set, to evaluate its influence on hidden state recovery and prediction accuracy. On the other hand, we simulated unreliable observed states in training set, and evaluated the influence of uncertain labelling on hidden state inference and prediction accuracy.

This section starts with the description of the protocol used to simulate data in both experimental settings. Then, the section focuses on implementation aspects. We next present and discuss the results obtained in both experimental settings.

\subsection{Simulated datasets}

This subsection first focuses on the model used to generate data. Then we describe the precursor sets used to further generate the test-set and the training datasets. 

\subsubsection{Generative model}
\label{generative_model}

These experiments were achieved on simulated data from a 4-state PHMC-LAR($2$) model whose transition matrix and initial probabilities are:

\begin{equation}
\label{4.regime.mc}
A = \begin{pmatrix}
		0.5 & \, 0.2 & \, 0.1 & \, 0.2 \\
		0.2 & \, 0.5 & \, 0.2 & \, 0.1 \\
		0.1 & \, 0.2 & \, 0.5 & \, 0.2 \\
		0.2 & \, 0.1 & \, 0.2 & \, 0.5
	\end{pmatrix}, \quad
	\pi = (0.25, \, 0.25, \, 0.25, \, 0.25).
\end{equation}

Within each state $k \in \{1, 2, 3, 4\}$, the autoregressive dynamics is a LAR($2$) process defined by parameters $\bm{\theta}^{(X,k)} = (\phi_{0,k}, \, \phi_{1,k}, \,\phi_{2,k}, \, h_k)$:

\begin{equation}
\label{lar.proc}
\begin{split}
\bm{\theta}^{(X,1)} &=  (2, \, 0.5, \, 0.75, 0.2), \quad \bm{\theta}^{(X,2)} =  (-2, -0.5, \, 0.75, \, 0.5), \\
\bm{\theta}^{(X,3)} &=  (4, \, 0.5, -0.75, \, 0.7), \quad \bm{\theta}^{(X,4)} =  (-4, -0.5, -0.75, \, 0.9).
\end{split}
\end{equation}

In the LAR(2) process associated with state $k$, stationarity is guaranteed by setting the following contraints: $\phi_{i,k} < 1,\ i \in \{1,2\}$.

Finally, the initial law $g_0$ is a bivariate Gaussian distribution\\

\begin{equation}
\label{init.law}
 g_0  = \mathcal{N}_2 \, \left((3, 5), \, \begin{pmatrix}
                                             1 & \, 0.1 \\ 0.1 & \, 1
 					  \end{pmatrix} \right).
\end{equation}

\begin{figure}[t]
\centering
    \begin{subfigure}[b]{0.45\textwidth}
        \includegraphics[width=\textwidth, trim=1cm 0cm 2cm 1cm, clip]{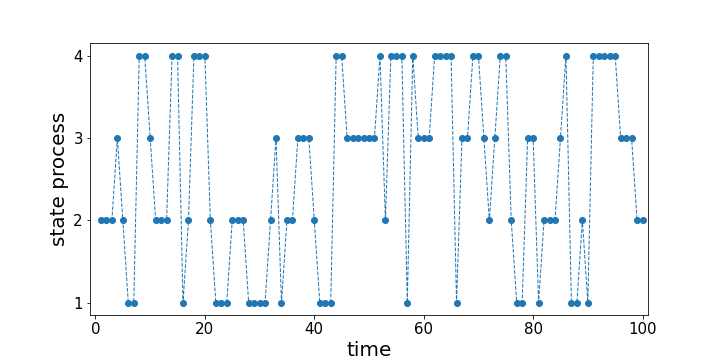}
        \caption{}
        \label{phmclar.state.process}
	\end{subfigure}
     \begin{subfigure}[b]{0.45\textwidth}
        \includegraphics[width=\textwidth, trim=1cm 0cm 2cm 1cm, clip]{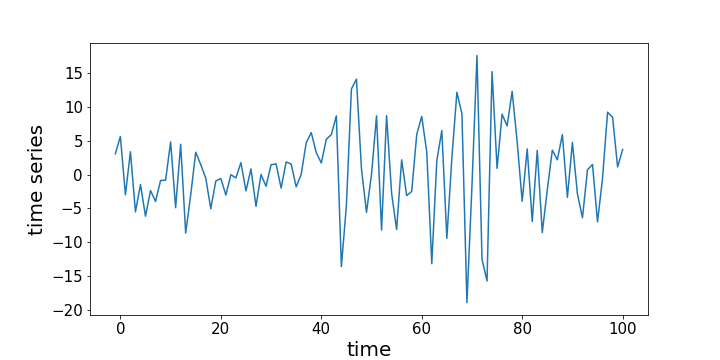}
        \caption{}
        \label{phmclar.time.series}
    \end{subfigure} 
\caption{A simulation from the PHMC-LAR($2$) model defined by Eq. \ref{4.regime.mc}-\ref{init.law}: (a) state process, (b) the corresponding time series}
\label{example.of.simulated.phmclar}
\end{figure}

Figure \ref{example.of.simulated.phmclar} shows an example of state process (Fig. \ref{phmclar.state.process}) and corresponding time series (Fig. \ref{phmclar.time.series}) that were simulated from the previously defined PHMC-LAR($2$). 

\subsubsection{Precursor sets for the test-set and training datasets}
\label{precursor_sets}
The training and test sets are common to both experimental settings (influence of the percentage of observed labels, influence of labelling error).
\paragraph{Inference}
The precursor set $\mathcal{P}_{infer\_test}$ of the test-set is composed of $M=100$ fully labelled observation sequences of length $\ell= 1000$. These sequences were generated from the PHMC-LAR(2) model described in Eq. \ref{4.regime.mc}-\ref{init.law}.
A protocol repeated for each $N \in \{1, 10, 100\}$ produced a precursor set $\mathcal{P}_{N\_infer\_train}$ consisting of $N$ fully labelled observation sequences of length $T=100$. The generative model in Eq. \ref{4.regime.mc}-\ref{init.law} was used for this purpose.
\paragraph{Forecasting}
In this case, training sets are each reduced to a single sequence. In each such sequence, the sequence's prefix of size $T=100$ is used for model training, whereas the subsequence $T+1, \cdots, T+10$ is used for testing prediction accuracy. The sequences of the unique precursor set denoted $\mathcal{P}_{N=1\_forecast\_train\_test}$ are generated using Eq. \ref{4.regime.mc}-\ref{init.law}. 
\subsection{Implementation}
\label{implementation}

Our experiments required intensive computing resources from a Tier 2 data centre (Intel 2630v4, 2$\times$10 cores 2.2 Ghz, 20$\times$6 GB). We exploited data-driven parallelization to replicate our experiments on various training sets. On the other hand, code parallelization allowed us to process multiple sequences simultaneously in the step E of the EM algorithm. The software programs dedicated to model training, hidden state inference and forecasting were written in Python 3.6.9. We used the NumPy and Scipy Python libraries.

The models were learnt through the EM algorithm with precision $\kappa = 10^{-6}$ and initialization procedure parameters $(L, N_{iter}) = (5, 10)$.

\subsection{Influence of the percentage of observed states}
\label{exp.1}

To analyze the impact of observed states, we varied the percentage $P$ of labelled observations (equivalently the percentage of observed states) in the training sets. $P$ was varied from $0\%$ (fully unsupervised case) to $100\%$ (fully supervised case), with steps of $10\%$. The aim is to evaluate the performance of intermediate cases for different sizes of the training datasets.

\subsubsection{Hidden state inference}
\label{exp1.eval.inference}

The test-set $\mathcal{S}_{infer\_test}$ was generated by unlabelling all states from the precursor set $\mathcal{P}_{infer\_test}$ described in Subsection \ref{precursor_sets} ($M=100$ fully observed sequences of length $\ell=1000$).

To generate the training sets, the following protocol was repeated for each $N \in \{1, 10, 100\}$ and for each percentage $P$: (i) considering the appropriate precursor set $\mathcal{P}_{N\_infer\_train}$ ($N$ fully observed sequences of length $T=100$) depicted in Subsection \ref{precursor_sets}, only a proportion of $P$ observations was kept labelled while the rest was unlabelled; (ii) this process was repeated $15$ times, each time varying which observations are kept labelled. Thus were produced $15$ training datasets $\mathcal{S}_{N\_P\_infer\_train\_1}, \cdots, \mathcal{S}_{N\_P\_infer\_train\_15}$.

The PHMC-LAR(2) model with 4 states was trained on each training set $\mathcal{S}_{N,P,infer\_train\_i}$, $i=1, \cdots, 15$. For each trained model, state inference was achieved for the $M$ fully hidden sequences of test-set $\mathcal{S}_{infer\_test}$, which yielded $M$ sequences of predicted labels. Inference performance was evaluated by comparing the true state sequences with the inferred ones, using the \textbf{Mean Percentage Error} (MPE) score defined as follows:

\begin{equation}
	\text{MPE} = \frac{1}{M} \sum_{i=1}^{M} \left[ \frac{1}{\ell} \sum_{j=1}^{\ell} \bm{1}_{s_j \neq \hat{s}_j} \right],
\end{equation}

where $s_j$'s and $\hat{s}_j$'s are respectively observed and inferred states. The MPE score varies between $0$ and $1$. The lower the value of the MPE score, the higher the inference performance.

\begin{figure}[t]
\centering
    \begin{subfigure}[b]{0.45\textwidth}
        \includegraphics[width=\textwidth, trim=1cm 0cm 2cm 1cm, clip]{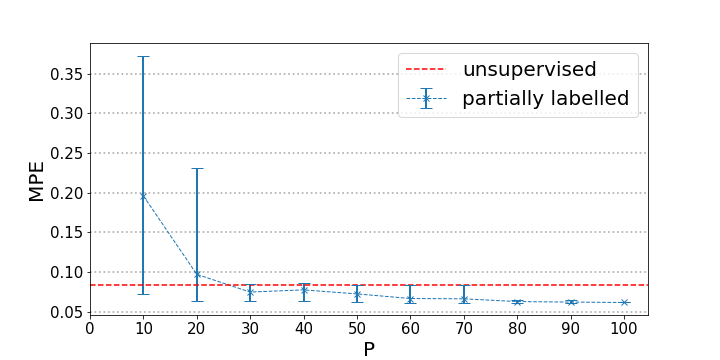}
        \caption{N = 1}
        \label{inference.exp1.N.1}
    \end{subfigure} 
    \begin{subfigure}[b]{0.45\textwidth}
        \includegraphics[width=\textwidth, trim=1cm 0cm 2cm 1cm, clip]{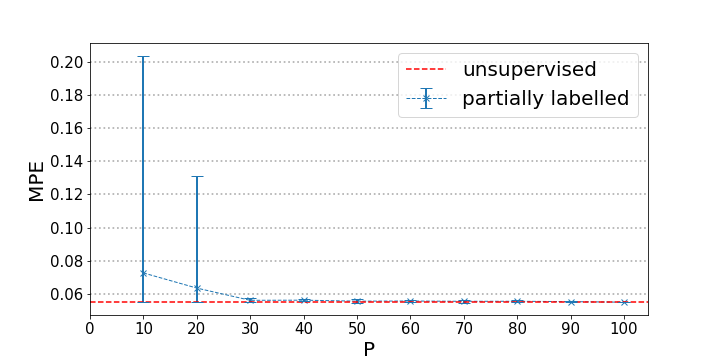}
        \caption{N = 10}
        \label{inference.exp1.N.10}
	\end{subfigure}
    \begin{subfigure}[b]{0.45\textwidth}
        \includegraphics[width=\textwidth, trim=1cm 0cm 2cm 1cm, clip]{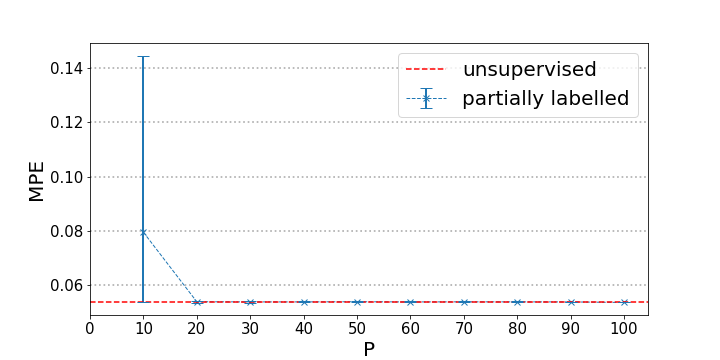}
        \caption{N = 100}
        \label{inference.exp1.N.100}
	\end{subfigure}
\caption{$95 \%$ confidence interval for mean percentage error (MPE) of hidden state inference, as a function of the percentage $P$ of labelled observations. Models were trained on datasets of $N$ sequences of length $100$, for each of $15$ replicates differing by the $P \%$ labelled observations. For each model, inference was performed for a test-set of $100$ unlabelled sequences of size $1000$. The $95 \%$ confidence interval of the MPE score was computed from the $15$ replicates. The dash (red) line indicates the MPE score obtained for the unsupervised learning case ($P = 0\%$). Mind the differences in scales between the three subfigures.}
\label{inference.exp1}
\end{figure}

Figure \ref{inference.exp1} displays $95\%$ confidence interval for the MPE score as a function of $P$. 
As expected, the results show that inference ability increases with the number of training sequences denoted by $N$. Note that when the proportion of labelled observations is less than some threshold ($P=30\%$ for $N=1, 10$ and $P=20\%$ for $N=100$), inference performance is greatly impacted by the distribution of observed states since we obtain very large confidence intervals for the MPE score.

\begin{figure}[t]
\centering
    \begin{subfigure}[b]{0.45\textwidth}
        \includegraphics[width=\textwidth, trim=1cm 0cm 2cm 1cm, clip]{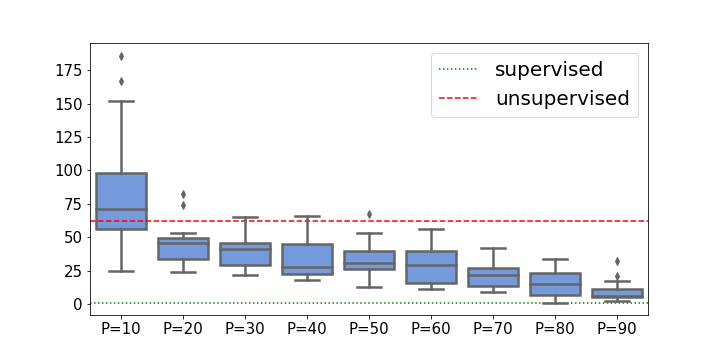}
        \caption{N = 1}
        \label{nbiter.exp1.N.1}
    \end{subfigure} 
    \begin{subfigure}[b]{0.45\textwidth}
        \includegraphics[width=\textwidth, trim=1cm 0cm 2cm 1cm, clip]{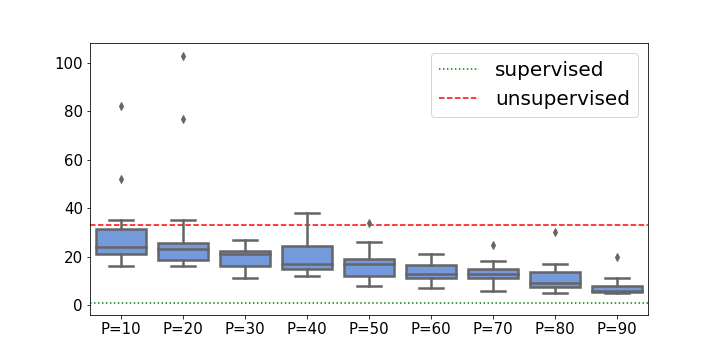}
        \caption{N = 10}
        \label{nbiter.exp1.N.10}
	\end{subfigure}
    \begin{subfigure}[b]{0.45\textwidth}
        \includegraphics[width=\textwidth, trim=1cm 0cm 2cm 1cm, clip]{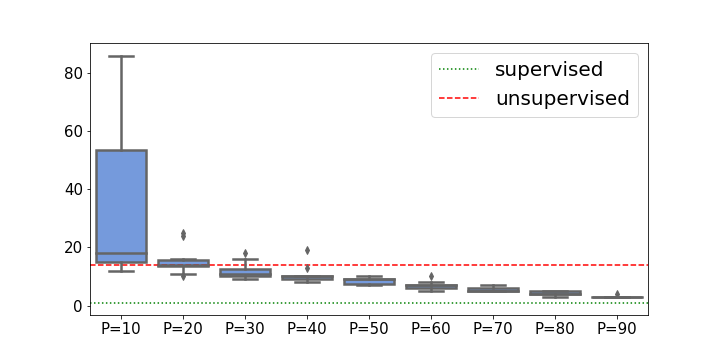}
        \caption{N = 100}
        \label{nbiter.exp1.N.100}
	\end{subfigure}
\caption{Number of EM iterations before convergence as a function of the percentage $P$ of labelled observations. For the description of the experimental protocol, see caption of Fig.\ref{inference.exp1}. The distribution of the number of EM iterations is studied across $15$ replicates. Dash (red) line and dot (green) line indicate the number of iterations for unsupervised and supersived learning cases respectively. Mind the differences in scales between the three subfigures.} 
\label{nbiter.exp1}
\end{figure}

\begin{figure}[h!]
\centering
    \begin{subfigure}[b]{0.45\textwidth}
        \includegraphics[width=\textwidth, trim=1cm 0cm 0cm 0cm, clip]{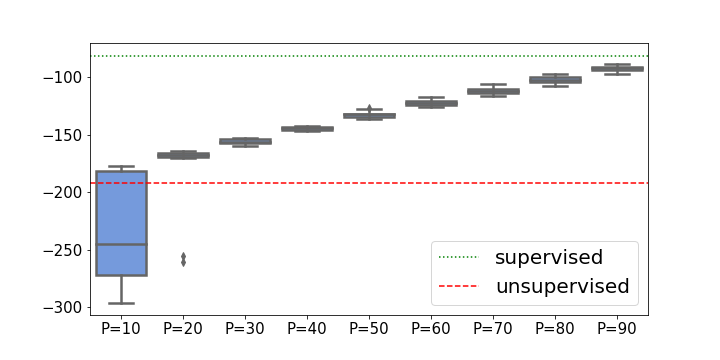}
        \caption{N = 1}
        \label{log.ll.exp1.N.1}
    \end{subfigure} 
    \begin{subfigure}[b]{0.45\textwidth}
        \includegraphics[width=\textwidth, trim=1cm 0cm 0cm 0cm, clip]{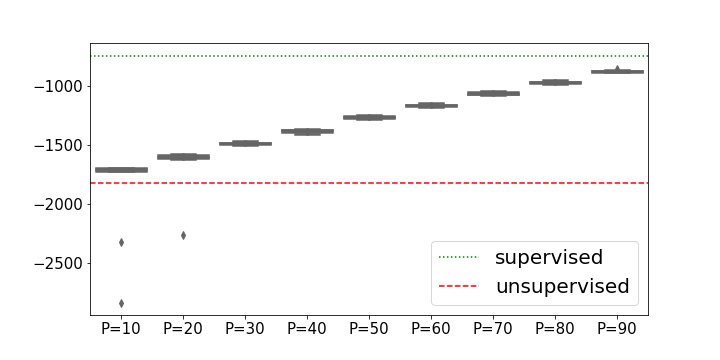}
        \caption{N = 10}
        \label{log.ll.exp1.N.10}
	\end{subfigure}
    \begin{subfigure}[b]{0.45\textwidth}
        \includegraphics[width=\textwidth, trim=1cm 0cm 0cm 0cm, clip]{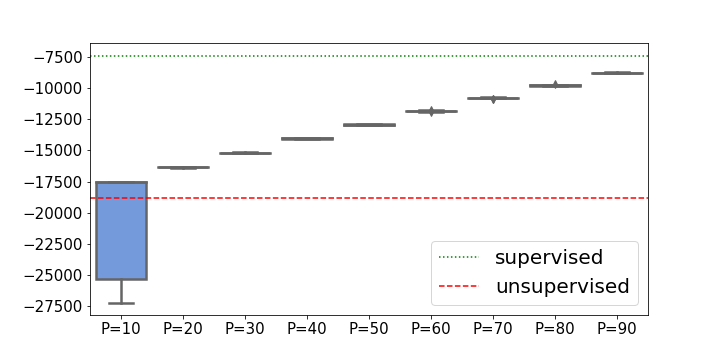}
        \caption{N = 100}
        \label{log.ll.exp1.N.100}
	\end{subfigure}
\vspace{-2mm}
\caption{Log-likelihood as a function of the percentage $P$ of labelled observations. For the description of the experimental protocol, see caption of Fig.\ref{inference.exp1}. The distribution of the log-likelihood is studied across $15$ replicates. Dash (red) line and dot (green) line indicate the log-likelihoods for unsupervised and supervised learning cases respectively. Mind the differences in scales between the three subfigures.}
\label{log.ll.exp1}
\end{figure}

For $N = 1$, the use of labelled observations makes it possible to outperform the fully unsupervised case ($P = 0\%$) (which translates into small MPE scores) when at least $30\%$ of observations are labelled (see Fig. \ref{inference.exp1.N.1}). In contrast, for $N=10, 100$, from some threshold value of $P$ (respectively $30\%$ and $20\%$), the use of larger proportions of labelled observations sustains inference performances equal to that of the fully unsupervised case (see Fig. \ref{inference.exp1.N.10} and \ref{inference.exp1.N.100}). Importantly, the results show that using large proportions of labelled observations considerably speeds up model training by decreasing the number of iterations of the EM algorithm (see Fig. \ref{nbiter.exp1}), and allows to better characterize the training data (which is reflected by a greater likelihood, see Fig. \ref{log.ll.exp1}). \cite*{ramasso-denoeux_2013_jour-ieee-transact-fuzzy-sys_partially-hmms} had already underlined the beneficial impact of partial knowledge integration on EM convergence in HPMCs. Our work confirms this advantage in the PHMC-LAR model, with a good preservation of inference performance.


\begin{figure}[t]
\centering
    \begin{subfigure}[b]{0.45\textwidth}
        \includegraphics[width=\textwidth, trim=0cm 0cm 2cm 1cm, clip]{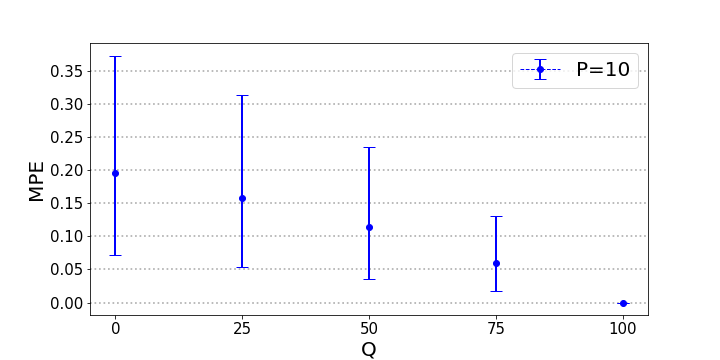}
        \caption{N = 1}
        \label{inference.exp1.bis.N.1}
    \end{subfigure} 
    \begin{subfigure}[b]{0.45\textwidth}
        \includegraphics[width=\textwidth, trim=0cm 0cm 2cm 1cm, clip]{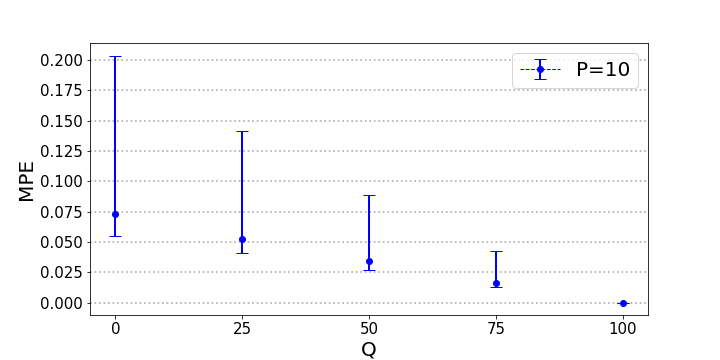}
        \caption{N = 10}
        \label{inference.exp1.bis.N.10}
	\end{subfigure}
    \begin{subfigure}[b]{0.45\textwidth}
        \includegraphics[width=\textwidth, trim=0cm 0cm 2cm 1cm, clip]{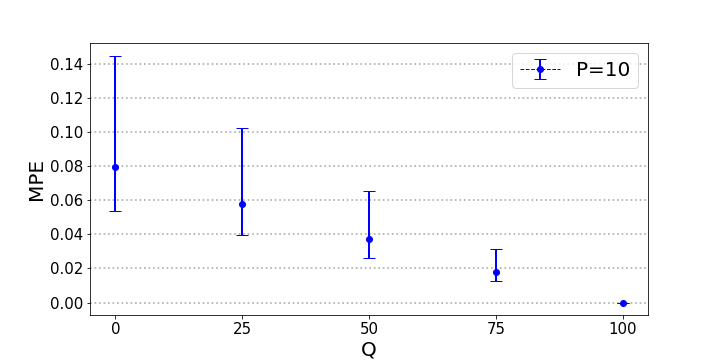}
        \caption{N = 100}
        \label{inference.exp1.bis.N.100}
	\end{subfigure}
\caption{$95 \%$ confidence interval for mean percentage error (MPE) of hidden state inference, as a function of the percentage $Q$ of labelled observations within test-set, with $P = 10\%$ labelled observations in the training sets. Models were trained on datasets of $N$ sequences of length $100$ in which $P=10\%$ of observations have been labelled. Fifteen replicates differing by the $P=10 \%$ labelled observations were considered. For each model, inference was performed for a test-set of $100$ partially labelled ($Q\%$) sequences of size $1000$. {\bf The $95 \%$ confidence interval of the MPE score was computed from the $15$ replicates.} Mind the differences in scales between the three subfigures.}
\label{inference.exp1.bis}
\end{figure}

In order to evaluate the influence of observed states in recognition phase, we considered the case $P=10 \%$ which previously obtained the lowest inference performance. This time, we also kept labelled a proportion $Q$ of observations within the test-set $\mathcal{S}_{infer\_test}$. We assessed the inference performances for the models trained on $\mathcal{S}_{N,P=10\%,infer\_train\_i}$, $i=1, \cdots 15$. Figure \ref{inference.exp1.bis} presents MPEs as a function of $Q$ for $N = 1, 10$ and $100$. We observe that inference performances are improved by the presence of observed states. 
More precisely, for $Q$ taking its values in $25 \%$, $50 \%$ and $75 \%$, respectively, MPE decreases by: (i) $19\%$, $42\%$ and $69\%$ for $N=1$ (Fig. \ref{inference.exp1.bis.N.1}); (ii) $27\%$, $52\%$ and $77\%$ for $N=10$ (Fig. \ref{inference.exp1.bis.N.10}); and (iii) $27\%$, $53\%$ and $77\%$ for $N=100$. (Fig. \ref{inference.exp1.bis.N.100}). These results show the ability of our variant of the Viterbi algorithm to infer partially-labelled sequences.

%
%
\subsubsection{Forecasting}
\label{exp1.eval.forecasting}
In this experiment, we consider models trained on a single sequence. This case corresponds to many real-world situations in which a unique time series is available ({\it e.g.}, the evolution of air pollution at some geographical location). Using the precursor set $\mathcal{P}_{N=1\_forecast\_train\_test}$ described in Subsection \ref{precursor_sets}, we generated datasets $\mathcal{S}_{N=1\_forecast\_train\_test\_i}$, $i=1, \cdots, 15$ each composed of a single sequence of size $110$. Again, the $15$ replicates differed by the $P\%$ labelled observations. In these sets, the sequence prefixes of length $T=100$ were used to train the models. \textit{Out-of-sample} forecasting was carried out at horizons $T+h$, $h=1, \dots, 10$, which means that prediction accuracy was assessed using subsequences $T+1, \cdots, T+h$. To note, the $P\%$ labelled observations were distributed in the sequence prefixes of length $T$.

Two experimental schemes were considered. First, the states at forecast horizons were supposed to be latent; that is, all states were unlabelled from $T+1$ to $T+h$, $h=1, \cdots, 10$. Then, we performed the prediction evaluation when states are observed at forecast horizons. The latter situation corresponds to performing the prediction conditional on some assumption on the regime. For instance, in econometrics, assuming we know which phase will be on (growth phase {\it versus} recession) might improve the forecasting performance of the Gross National Product (GNP). In this case, all states were kept labelled from $T+1$ to $T+h$, $h=1, \cdots, 10$. 

Prediction performance is estimated by the \textbf{Root Mean Square Error} (RMSE) defined as follows:
  
%

%
\begin{equation}
	\text{RMSE}_h = \sqrt{\frac{1}{N_{rep}} \sum_{i=1}^{N_{rep}} (X_{T+h}^{(i)} - \hat{X}_{T+h}^{(i)})^2},
\end{equation}

where $h$ is the forecast horizon and $N_{rep}=15$ is the number of replicates. Accurate predictions are characterized by low RMSEs. 
 
\begin{figure}[]
\centering
\begin{subfigure}[b]{0.45\textwidth}
	\includegraphics[width=\textwidth, trim=1cm 0cm 2cm 1cm, clip]{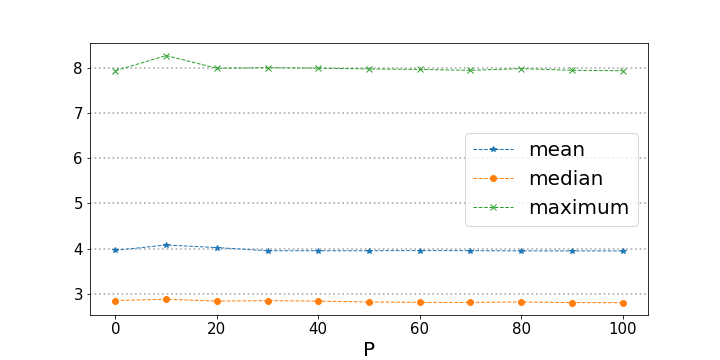}
	\caption{}
	\label{forecasting.exp1.dist.unknown}
\end{subfigure}
\begin{subfigure}[b]{0.45\textwidth}  
	\includegraphics[width=\textwidth, trim=1cm 0cm 2cm 1cm, clip]{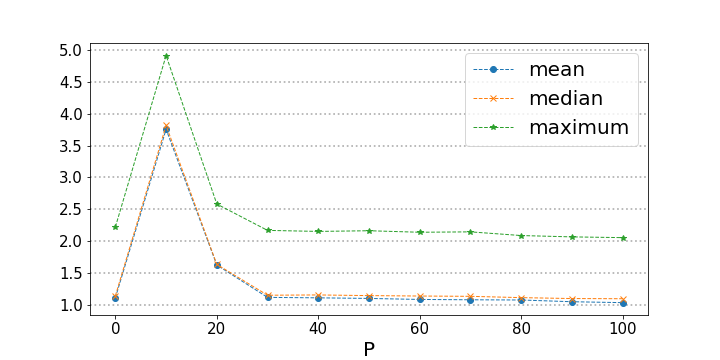}
	\caption{}
	\label{forecasting.exp1.dist.known}
\end{subfigure}
\caption{Mean, median and maximum root mean square error (RMSE) of prediction at horizon $h$ as a function of $P$, the percentage of labelled observations in the training datasets. States at forecast time-steps $T+h,$ $h=1, \cdots 10$ are (a) hidden, (b) known. Models were trained on a single sequence, for each of $15$ replicates differing by the $P \%$ labelled observations. Model training was performed on subsequences of length $100$, whereas prediction was achieved for the $10$ subsequent time-steps. For each value of $P$, the statistics provided were computed across the $15$ replicates and all horizons.}
\label{forecasting.exp1.dist}
\end{figure}

Table \ref{exp1.prediction.perf} presents the RMSEs obtained when the states at forecast horizons are supposed to be latent. Fig. \ref{forecasting.exp1.dist.unknown} presents the mean, median and maximum of RMSEs, computed over all forecast horizons, as a function of $P$, the percentage of labelled observations in the training sets. Table \ref{exp1.prediction.perf} and Fig. \ref{forecasting.exp1.dist.unknown} show that as from some low $P$ threshold ($10\%$ or $20\%$), the prediction performance remains nearby constant across proportions.

%
\begin{table}[t!]
\scalefont{0.85}
\centering
\begin{tabular}{p{0.5cm} p{0.6cm} p{0.6cm} p{0.6cm} p{0.6cm} p{0.6cm} p{0.6cm} p{0.6cm} p{0.6cm} p{0.6cm} p{0.6cm}}
\hline
\backslashbox{$P$}{h} & 1 & 2 & 3 & 4 & 5 & 6 & 7 & 8 & 9 & 10 \\
\hline
0  & 1.860 & \bf{6.680} & 1.830 & 3.165 & 4.167 & 2.540 & 1.133 & 7.938 & 7.854 & 2.465 \\
10 & 2.035 & 8.273 & 1.829 & \bf{2.909} & 4.477 & 2.851 & \bf{0.957} & \bf{7.667} & \bf{7.583} & \bf{2.224}  \\
20 & 1.934 & 7.612 & \bf{1.337} & 3.161 & 4.110 & 2.482 & 1.189 & 7.991 & 7.907 & 2.518 \\
30 & 1.323 & 7.450 & 1.373 & 3.168 & \bf{4.093} & \bf{2.469} & 1.201 & 8.005 & 7.921 & 2.532  \\
40 & 1.293 & 7.496 & 1.392 & 3.158 & 4.103 & 2.480 & 1.191 & 7.994 & 7.911 & 2.521 \\
50 & 1.308 & 7.525 & 1.402 & 3.135 & 4.122 & 2.496 & 1.174 & 7.978 & 7.894 & 2.505 \\
60 & 1.394 & 7.502 & 1.424 & 3.115 & 4.134 & 2.508 & 1.162 & 7.965 & 7.882 & 2.493 \\
70 & 1.363 & 7.560 & 1.431 & 3.094 & 4.155 & 2.527 & 1.142 & 7.946 & 7.862 & 2.473 \\
80 & 1.306 & 7.502 & 1.395 & 3.129 & 4.116 & 2.489 & 1.179 & 7.984 & 7.900 & 2.511 \\
90 & 1.294 & 7.569 & 1.444 & 3.088 & 4.155 & 2.526 & 1.142 & 7.947 & 7.863 & 2.473  \\
100 & \bf{1.267} & 7.613 & 1.447 & 3.076 & 4.164 & 2.535 & 1.132 & 7.937 & 7.854 & 2.464 \\
\hline
\end{tabular}
\caption{Root mean square error (RMSE) of prediction at horizon $h$ for different values of $P$, when the states are unknown throughout forecast horizons. $P$ is the percentage of labelled observations within the training datasets. The forecast horizons are time-steps $T+1$ to $T+h$, $T=100$. For a given value of $P$, models were each trained on a unique sequence: the sequence's prefix of length $T=100$ was used for training, for each of $15$ replicates differing by the $P \%$ labelled observations distributed in the prefix. Then, out-of-sample forecasting was carried out at time-steps $T+1, \dots, T+10$, for the same sequence. The figures in bold highlight the minimum RMSE obtained across all labelling percentages ($P$), at each horizon ($h$) considered}
\label{exp1.prediction.perf}
\scalefont{1.0}
\end{table}

In addition, Table \ref{exp1.prediction.perf} also highlights that the ability to predict depends on the forecast horizon under consideration. At any given labelling percentage $P$, high RMSE scores ({\it i.e.}, around $7$) alternate with low scores (around $1$) across horizons. The nonmonotonic error trend across horizons was observed empirically for MS-AR models and threshold autoregressive models when they are applied to US GNP time series \citep{clements-krolzig_1998_journ-econometrics_forecasting-msar-vs-threshold-ar}.

Finally, our experiments show that PHMC-LAR model's ability to better characterize the training data in presence of large proportions of labelled observations (characterized by greater likelihood, see Fig. \ref{log.ll.exp1.N.1}) does not translate into an improved forecast performance.

When states are known at forecast horizons, RMSEs (presented in Table \ref{exp1.prediction.perf.observed.states}) are reduced by $44\%$ on average. Moreover, Fig. \ref{forecasting.exp1.dist.known} shows that above percentage $P = 30 \%$, prediction performances are slightly greater than that of the unsupervised case ($P=0 \%$). Note that as in the case when the states are unknown at forecast horizons, the prediction ability depends on the forecast horizon. Again, for a given $P$, the RMSE score does not systematically increase with forecast horizon $h$, although previously predicted values are used as inputs when predicting at next horizons.

\begin{table}[ht]
\scalefont{0.85}
\centering
\begin{tabular}{p{0.5cm} p{0.6cm} p{0.6cm} p{0.6cm} p{0.6cm} p{0.6cm} p{0.6cm} p{0.6cm} p{0.6cm} p{0.6cm} p{0.6cm} }
\hline
\backslashbox{$P$}{h} & 1 & 2 & 3 & 4 & 5 & 6 & 7 & 8 & 9 & 10 \\
\hline
0  & 0.083 & 0.325 & 0.870 & 1.577 & 1.509 & 1.171 & 2.220 & 1.216 & \bf{0.996} & \bf{1.104} \\
10 & 3.730 & 1.791 & 2.936 & 3.593 & 4.914 & 4.153 & 4.060 & 4.873 & 3.603 & 3.902 \\
20 & 1.831 & 0.510 & 1.806 & 1.939 & 2.171 & 1.250 & 2.581 & 1.438 & 1.239 & 1.458  \\
30 & 0.083 & \bf{0.321} & 0.854 & 1.542 & 1.477 & 1.158 & 2.167 & 1.137 & 1.109 & 1.289  \\
40 & 0.070 & 0.325 & 0.841 & 1.532 & 1.460 & 1.150 & 2.151 & 1.154 & 1.084 & 1.301 \\
50 & 0.065 & 0.324 & 0.832 & 1.540 & 1.459 & 1.154 & 2.161 & 1.130 & 1.078 & 1.229  \\
60 & 0.063 & 0.329 & 0.829 & 1.524 & 1.443 & 1.145 & 2.137 & 1.125 & 1.039 & 1.181  \\
70 & 0.057 & 0.329 & 0.810 & 1.531 & 1.431 & 1.143 & 2.143 & 1.118 & 1.036 & 1.263   \\
80 & 0.036 & 0.327 & 0.810 & 1.490 & 1.411 & 1.134 & 2.086 & 1.086 & 1.072 & 1.276 \\
90 & 0.036 & 0.325 & 0.788 & 1.479 & 1.386 & 1.124 & 2.065 & 1.067 & 1.023 & 1.161  \\
100 & \bf{0.001} & 0.326 & \bf{0.760} & \bf{1.473} & \bf{1.368} & \bf{1.121} & \bf{2.053} & \bf{1.065} & 1.002 & 1.133 \\
\hline
\end{tabular}
\caption{Root mean square error (RMSE) of prediction at horizon $h$ for different values of $P$, when the states are known throughout forecast horizons. $P$ is the percentage of labelled observations within the training datasets. The forecast horizons are time-steps $T+1$ to $T+h$, $T=100$. For the description of the experimental protocol, see caption of Table \ref{exp1.prediction.perf}. The states are known from $T+1$ to $T+10$ time-steps. The figures in bold highlight the minimum RMSE obtained across all labelling percentages ($P$), at each horizon ($h$) considered}
\label{exp1.prediction.perf.observed.states}
\scalefont{1.0}
\end{table}
\subsection{Influence of labelling error}
\label{exp.2}
In this experiment, the influence of labelling error is evaluated. To simulate unreliable labels, we proceeded as follows. 

At each time-step $t$, an error probability  $p_t$ was drawn randomly from a beta distribution with mean $\rho$ and variance $0.2$. With probability $p_t$, the observed state $s_t$ was replaced by a random state uniformly chosen from $\{1, 2, 3, 4\} \setminus \{s_t\}$. So, the unreliable labels $\tilde{s}_t$ were defined as follows:

\begin{equation}
\label{unreliable.labels}
\begin{split}
&p_t \sim \beta(0.2,\rho)\\
&\tilde{s}_t  = 
 	\left\{ 
      \begin{aligned}
         s_t  \qquad \qquad \qquad \text{with probability} \quad   1 - p_t  \\
         \mathcal{U} \left(\{1, 2, 3, 4\} \setminus \{s_t\} \right) \quad \text{with probability}  \quad p_t
      \end{aligned}
    \right.
\end{split}
\end{equation}

where $\mathcal{U}$ is the discrete-valued uniform distribution. Thus, on average a proportion $\rho$ of observations is assigned wrong labels. 

\subsubsection{Inference of hidden states}
\label{eval.inference}
To assess inference performance in presence of labelling errors, we relied on the test-set $\mathcal{S}_{infer\_test}$ described in Subsection \ref{exp.1} ($M=100$ fully hidden sequences of length $\ell=1000$) corresponding to the fully labelled dataset $\mathcal{P}_{infer\_test}$.

To generate the training sets, for each N $\in \{1,\ 10,\ 100\}$, we considered the appropriate precursor set $\mathcal{P}_{N\_infer\_train}$ ($N$ fully observed sequences of length $T=100$) depicted in Subsection \ref{precursor_sets}.

We varied the mean labelling error probability $\rho$ in $\{0.1, 0.2, 0.3, 0.4, 0.5, 0.6, 0.7,$ $0.8, 0.9, 0.95 \}$. For $N \in \{1,\ 10,\ 100\}$, and each value of $\rho$, we generated $15$ replicates from dataset $\mathcal{P}_{N\_infer\_train}$, each time varying the distribution of the wrong labels amongst the observations. The PHMC-LAR(2) model with 4 states was trained on each of the training sets $\mathcal{S}_{N,\rho,infer\_train\_1}, \cdots, \mathcal{S}_{N\_\rho\_infer\_train\_15}$ thus obtained.

For each trained model, state inference was achieved, which yielded $M=100$ sequences of predicted labels of length $1000$, to be compared with the label sequences within $\mathcal{P}_{infer\_test}$ (see Subsection \ref{generative_model}).

\begin{figure}[t]
\centering
    \begin{subfigure}[b]{0.45\textwidth}
        \includegraphics[width=\textwidth, trim=1cm 0cm 2cm 1cm, clip]{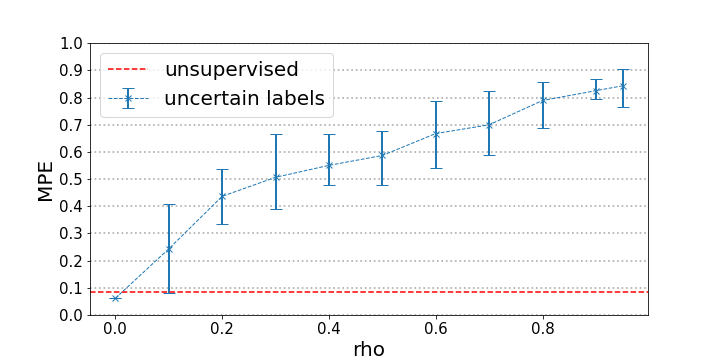}
        \caption{N = 1}
        \label{inference.exp2.N.1}
    \end{subfigure} 
    \begin{subfigure}[b]{0.45\textwidth}
        \includegraphics[width=\textwidth, trim=1cm 0cm 2cm 1cm, clip]{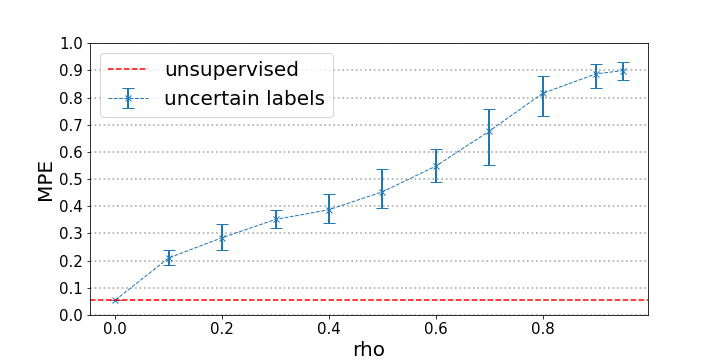}
        \caption{N = 10}
        \label{inference.exp2.N.10}
	\end{subfigure}
    \begin{subfigure}[b]{0.45\textwidth}
        \includegraphics[width=\textwidth, trim=1cm 0cm 2cm 1cm, clip]{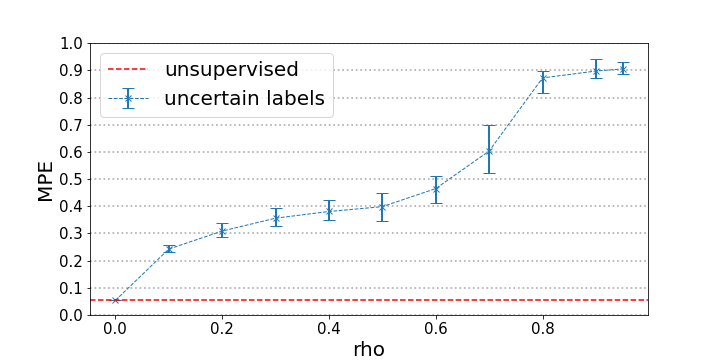}
        \caption{N = 100}
        \label{inference.exp2.N.100}
	\end{subfigure}
\caption{$95 \%$ confidence interval for mean percentage error (MPE) of hidden state inference, as a function of the mean labelling error probability $\rho$. Models were trained on $N$ sequences, for each of $15$ replicates differing by the $\rho \%$ ill-labelled observations. The average MPE was computed from the $15$ replicates. The dash (red) line indicates the MPE score obtained for the unsupervised learning case. Mind the differences in scales between the three subfigures.}
\label{inference.exp2}
\end{figure}

 Figure \ref{inference.exp2} presents $95\%$ confidence intervals for the MPE score as a function of $\rho$. Note that for all sizes $N \in \{1,\ 10,\ 100\}$ of training data, the average MPE gradually increases when $\rho$ tends to $1$. Moreover, confidence intervals become more and more tight when larger training data is considered. We also observe that up to $\rho = 0.7$, the robustness to labelling errors, translated into small MPE average and low dispersion, increases with $N$. However, from $\rho \ge 0.8$, this trend is reversed and inference performance slightly decreases when $N$ grows.

On the other hand, we underline that the fully unsupervised case outperforms supervised cases in presence of labelling errors. Up to relatively high labelling error rates ($\rho=70\%$), the trade-off between training time and inference performance becomes beneficial for large training datasets. For instance, for $N=100$, with a $70\%$-reliable labelling function ({\it i.e.} $\rho = 0.3$), the EM algorithm converges after a single iteration against $67$ iterations for the unsupervised case; and the resulting model has good inference abilities with an MPE score equal to $35\%$ on average (see Fig. \ref{inference.exp2.N.100}) against $5\%$ on average in the unsupervised case. Thus, when analyzing real-world data for which the number of states $K$ and auto-regressive order $p$ are unknown, model selection strategies can capitalize on such labelling functions in order to explore/prospect larger grids of values for the hyperparameters $K$ and $p$. 

%

\subsubsection{Forecasting}
\label{eval.forecasting}
As in Subsection \ref{exp1.eval.forecasting}, we considered models trained on a single sequence ($N = 1$). Again, for each value of the mean labelling error probability $\rho$, we used precursor set $\mathcal{P}_{N=1\_forecast\_train\_test}$ described in Subsection \ref{precursor_sets}, and we varied the distribution of wrong labels: $15$ replicates ({\it i.e.}, $15$ sequences of length $T=100$) were thus generated. Out-of-sample forecasting was carried out at horizons $T+h$, $h = 1,\cdots,10$.
  
\begin{figure}[h]
\centering
    \begin{subfigure}[b]{0.45\textwidth}
        \includegraphics[width=\textwidth, trim=1cm 0cm 2cm 1cm, clip]{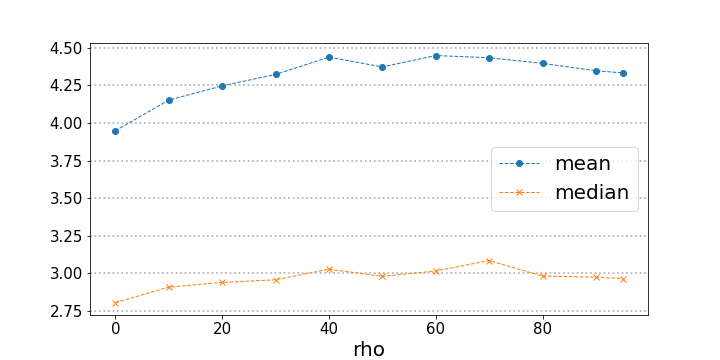}
        \caption{}
        \label{exp2.prediction.perf.mean}
	\end{subfigure}
	\begin{subfigure}[b]{0.45\textwidth}
        \includegraphics[width=\textwidth, trim=1cm 0cm 2cm 1cm, clip]{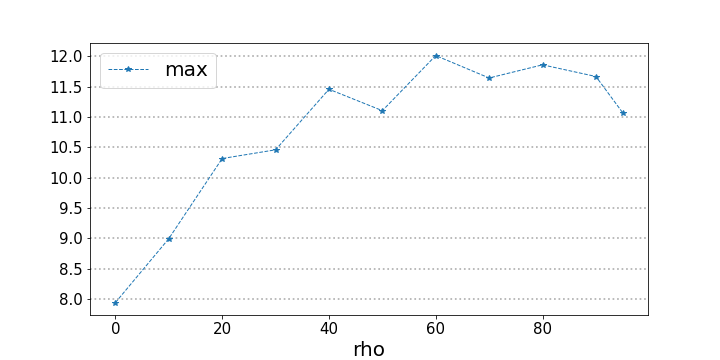}
        \caption{}
        \label{exp2.prediction.perf.max}
	\end{subfigure}
\caption{Descriptive statistics for the distribution of the root mean square error (RMSE) of prediction, as a function of $\rho$: (a) mean and median for prediction error, (b) maximum for prediction error. $\rho$ denotes the mean labelling error probability. The forecast horizons are time-steps $T+1$ to $T+h$, $T=100$. The statistics are computed over all horizons.}
\label{exp2.prediction.perf.fig}
\end{figure}

Table \ref{exp2.prediction.perf} presents RMSE scores for different values of mean labelling error $\rho$ when states are unknown at forecast horizons $h = 1, \dots, 10$. The results show that at forecast horizons $h = 1, 2, 5, 6$, the best prediction accuracies are reached when $\rho$ is null, whereas at the remaining horizons, the highest accuracies are obtained when $\rho = 0.8$ or $0.9$. Figure \ref{exp2.prediction.perf.fig} presents the mean, median and maximum for the prediction errors computed over the whole forecast horizons as a function of $\rho$. We observe that the mean and median very slightly increase with $\rho$, whereas labelling errors exert a greater impact on the maximum values of RMSEs. Therefore, this second experiment also highlights the remarkable robustness to error labelling in the prediction task, over the whole range of error rates.

\begin{table}[t]
\scalefont{0.85}
\centering
\begin{tabular}{p{0.5cm} p{0.6cm} p{0.6cm} p{0.6cm} p{0.6cm} p{0.6cm} p{0.6cm} p{0.6cm} p{0.6cm} p{0.6cm} p{0.6cm}}
\hline
\backslashbox{$\rho$}{h} & 1 & 2 & 3 & 4 & 5 & 6 & 7 & 8 & 9 & 10 \\
\hline
0   & \bf{1.267} & \bf{7.613} & 1.447 & 3.076 & \bf{4.164} & \bf{2.535} & 1.132 & 7.937 & 7.854 & 2.464 \\
0.1 & 1.814 & 8.992 & 1.393 & 3.193 & 4.334 & 2.625 & 1.117 & 7.865 & 7.780 & 2.407  \\
0.2 & 2.258 & 10.315 & 1.529 & 2.855 & 4.758 & 3.026 & 0.811 & 7.481 & 7.398 & 2.044 \\
0.3 & 2.793 & 10.458 & 1.575 & 2.911 & 4.689 & 3.004 & 0.801 & 7.512 & 7.426 & 2.062  \\
0.4 & 2.877 & 11.457 & 1.161 & 3.114 & 4.779 & 2.941 & 0.886 & 7.562 & 7.478 & 2.123  \\
0.5 & 2.655 & 11.104 & 1.396 & 2.953 & 4.812 & 3.008 & 0.843 & 7.488 & 7.410 & 2.062  \\
0.6 & 2.925 & 12.013 & 1.004 & 3.031 & 4.878 & 3.002 & 0.749 & 7.472 & 7.392 & 2.020  \\
0.7 & 3.088 & 11.643 & 1.271 & 2.969 & 4.901 & 3.082 & 0.706 & 7.409 & 7.321 & 1.954  \\
0.8 & 2.656 & 11.860 & \bf{0.905} & 3.001 & 4.848 & 2.954 & 0.768 & 7.498 & 7.422 & 2.046  \\
0.9 & 2.444 & 11.667 & 1.338 & \bf{2.786} & 5.011 & 3.164 & \bf{0.647} & \bf{7.310} & \bf{7.234} & \bf{1.875}  \\
0.95 & 2.362 & 11.071 & 1.192 & 3.027 & 4.685 & 2.905 & 0.866 & 7.588 & 7.504 & 2.135 \\
\hline
\end{tabular} 
\caption{Root mean square error (RMSE) of prediction at horizon $h$ for different values of the mean labelling error probability $\rho$, when the states are unknown throughout forecast horizons. The forecast horizons are time-steps $T+1$ to $T+h$, $T=100$. The states are unknown from $T+1$ to $T+10$ time-steps. For a given value of $\rho$, models were each trained on a unique sequence: the sequence's prefix of length $T=100$ was used for training, for each of $15$ replicates differing by the position of ill-labelled observations distributed in the prefix. Then, out-of-sample forecasting was carried out at time-steps $T+1, \dots, T+10$, for the same sequence. The figures in bold highlight the minimum RMSE obtained across all mean labelling error probabilities ($\rho$), at each horizon ($h$) considered}
\label{exp2.prediction.perf} 
\scalefont{1.0}
\end{table}

%

\newpage
\section{Conclusion}
\label{conclusion}
In this work, we have introduced the PHMC-LAR model to analyze time series subject to switches in regimes. Our model is a generalization of the well-known Hidden Regime-switching Autoregressive (HRSAR) and Observed Regime-switching Autoregressive (ORSAR) models when regime-switching is modelled by a Markov Chain. Our model allows to handle the intermediate case where the state process is partially observed.\\ 
\indent In the evaluation, we conducted our experiments on simulated data and considered both inference performance and prediction accuracy. The results show that the partially observed states (when they represent a reasonable proportion) allow a better characterization of training data (reflected by greater log-likelihood), in comparison with the unsupervised case. An interesting characteristics of the PHMC-LAR model is that the partially observed states allow faster convergence for the learning algorithm. This performance is obtained with no or practically no impact on the quality of hidden state inference, as from labelling percentages around $20\%$-$30\%$; the prediction accuracy is also preserved above such percentage thresholds. Furthermore, faster EM convergence is also verified in a fully supervised scheme where part of the observations is ill-labelled. Model selection strategies can therefore rely on an approximate labelling function (provided by an expert or by a supervised algorithm learnt on a small subset of data for which the true labels are known), to explore larger grids of hyperparameter values. In addition, complementary experimental studies have revealed the robustness of our model to labelling errors, particularly when large training datasets and moderate labelling error rates are considered. Finally, we showed the ability of our variant of the Viterbi algorithm to infer partially-labelled sequences.\\
\indent A natural extension of the PHMC-LAR model consists in putting uncertainty on partial knowledge: for instance instead of states observed with no doubt, a subset of possible states with various occurrence probabilities can be considered at each time-step.
On the other hand, it is more realistic to consider time-dependent state processes, especially when large time series are analyzed. These directions will be investigated in future work.

\section*{Acknowledgements}
The software development and the realization of the experiments were performed at the CCIPL (Centre
de Calcul Intensif des Pays de la Loire, Nantes, France).

\section*{Funding}
Fatoumata Dama is supported by a PhD scholarship granted by the French Ministery for Higher Education, Research and Innovation.

\appendix
\section{Appendix: backward-forward-backward algorithm}
\label{BFB}
The \textit{Backward-forward-backward} algorithm introduced by \cite*{scheffer-wrobel_2001_ecml-pkdd-workshop_partially-hidden-markov-models} for PHMC model learning has been adapted to the PHMC-LAR framework. This algorithm makes it possible to compute the probabilities 
\begin{align}
\label{def.xi.app}
	\xi_t(k, \ell) &= P(S_{t-1}=k, S_{t}=\ell \, | \, X_{1-p}^T=x_{1-p}^T, \Sigma; \, \hat{\bm{\theta}}), \quad \text{for} \quad t=2, \dots, T, \quad 1 \le k, \,\ell \le K
\end{align}
in $\mathcal{O}(TK^2)$ operations. The analytical development for the above quantity involves three additional probabilities:
\begin{align}
\label{compute.xi.b}
    \xi_t(k, \ell) 
    &= \frac{\beta_t(\ell) \, P(S_t=\ell \,|\, S_{t-1}=k; \, \hat{\bm{\theta}}) \, P(X_t=x_t \,|\, X_{t-p}^{t-1}, S_t=\ell; \, \hat{\bm{\theta}}) \, \alpha_{t-1}(k) \, \tau_t(\ell)} {P(X_1^T=x_1^T \,|\, X_{1-p}^0, \Sigma; \, \bm{\hat{\theta}}) \, \tau_{t-1}(k)}  \times \bm{1}_{\{\ell \in \sigma_t, \, k \in \sigma_{t-1}\}},
\end{align}
with
$$ \tau_t(s) = P(\sigma_{t+1}, \dots, \sigma_{T} \,|\, S_t=s, \hat{\bm{\theta}}), $$
$$ \alpha_t(s) = P(S_t=s, X_1^t=x_1^t \,|\, X_{1-p}^0, \Sigma; \,\hat{\bm{\theta}}), $$
$$ \beta_t (s) = P(X_{t+1}^T=x_{t+1}^T \,|\, X_{t+1-p}^t, S_t = s, \Sigma; \, \hat{\bm{\theta}}). $$
\noindent The algorithm operates recursively in three steps: two backward steps chained through a forward step. The first backward step computes the set of probabilities $\tau_t(s)$ (subsection \ref{bfb.first.backward.step}); the forward step computes probabilities $\alpha_t(s)$ (subsection \ref{bfb.forward.step}); the second backward step computes probabilities $\beta_t(s)$. 
In subsection \ref{BFB.scaling}, we describe a scaling method that is necessary to prevent floating point underflow when running the algorithm, especially when large sequences are considered.

\paragraph{\textbf{Proof.}} First, in Eq. \ref{proof.xi.2}, the conditional probability is transformed into a joint probability. Then, in Eq. \ref{proof.xi.3}, we successively maginalize $X_{t+1}^T$, $X_t$, $S_t$ and $(S_{t-1}, X_1^{t-1})$. According to the conditional independence graph of the PHMC-LAR model, the marginalization of $X_{t+1}^T$ gives $\beta_{t}(\ell)$, that of $X_t$ yields $P(X_t=x_t \, | \, S_{t}=\ell,  X_{t-p}^{t-1}, \Sigma; \, \hat{\bm{\theta}})$, that of $S_t$ gives $P(S_{t}=\ell \, | \, S_{t-1}=k, \Sigma; \, \hat{\bm{\theta}})$ and that of $(S_{t-1}, X_1^{t-1})$ provides $\alpha_{t-1}(k)$. Finally, in Eq. \ref{proof.xi.4}-\ref{proof.xi.6}, the probability $P(S_{t}=\ell \, | \, S_{t-1}=k, \Sigma; \, \hat{\bm{\theta}})$ is developed using Bayes' rule. Note that in Eq. \ref{proof.xi.6}, the probability $P(S_t=\ell, \sigma_t \,|\, S_{t-1}=k, \sigma_{t-1}; \, \hat{\bm{\theta}})$ is null for $\ell \notin \sigma_t$ and is not defined for $k \notin \sigma_{t-1}$.
\begin{align}
    \xi_t(k, \ell) 
    &= P(S_{t-1}=k, S_{t}=\ell \, | \, X_{1-p}^T=x_{1-p}^T, \Sigma; \, \hat{\bm{\theta}}) \nonumber \\
\label{proof.xi.2}
    &= \frac{P(S_{t-1}=k, S_{t}=\ell, X_1^T \, | \, X_{1-p}^0, \Sigma; \, \hat{\bm{\theta}})}{P(X_1^T \, | \, X_{1-p}^0, \Sigma; \, \hat{\bm{\theta}})} \\
\label{proof.xi.3}
    &= \quad P(X_{t+1}^T=x_{t+1}^T \, | \, S_{t-1}=k, S_{t}=\ell,  X_1^t,  X_{1-p}^0, \Sigma; \, \hat{\bm{\theta}}) \,  \nonumber \\
    & \quad \times P(X_t=x_t \, | \, S_{t-1}=k, S_{t}=\ell,  X_1^{t-1},  X_{1-p}^0, \Sigma; \, \hat{\bm{\theta}}) \,  \nonumber \\
    & \quad \times P(S_{t}=\ell \, | \, S_{t-1}=k,  X_1^{t-1},  X_{1-p}^0, \Sigma; \, \hat{\bm{\theta}}) \frac{P(S_{t-1}=k,  X_1^{t-1}=x_1^{t-1} \, | \, X_{1-p}^0, \Sigma; \, \hat{\bm{\theta}})}{P(X_1^T \, | \, X_{1-p}^0, \Sigma; \, \hat{\bm{\theta}})} \\
   &= \frac{\beta_t(\ell) \, P(X_t=x_t \, | \, S_{t}=\ell,  X_{t-p}^{t-1}, \Sigma; \, \hat{\bm{\theta}})   \, \alpha_{t-1}(k) }{P(X_1^T \, | \, X_{1-p}^0, \Sigma; \, \hat{\bm{\theta}})} \times P(S_{t}=\ell \, | \, S_{t-1}=k, \Sigma; \, \hat{\bm{\theta}})   
\end{align}
with 
\begin{align}
\label{proof.xi.4}
	P(S_{t}=\ell \, | \, S_{t-1}=k, \Sigma; \, \hat{\bm{\theta}}) 
	&= \frac{ P(S_{t}=\ell, S_{t-1}=k, \sigma_1^T; \, \hat{\bm{\theta}})}{P(S_{t-1}=k, \sigma_1^T; \, \hat{\bm{\theta}}) } \\
\label{proof.xi.5}
	&= \frac{ P(\sigma_{t+1}^T \,|\, S_{t}=\ell, S_{t-1}=k, \sigma_1^t; \, \hat{\bm{\theta}}) \, P(S_{t}=\ell, S_{t-1}=k, \sigma_1^t; \hat{\bm{\theta}} )} { P( \sigma_t^T \,|\, S_{t-1}=k, \sigma_1^{t-1}; \, \hat{\bm{\theta}}) \, P(S_{t-1}=k, \sigma_1^{t-1}; \, \hat{\bm{\theta}}) } \\
\label{proof.xi.6}
	&= \frac{\tau_{t}(\ell)}{\tau_{t-1}(k)} \times P(S_t=\ell, \sigma_t \,|\, S_{t-1}=k, \sigma_{t-1}; \, \hat{\bm{\theta}})  \\
	&= \left\{ 
	\begin{aligned}
		& \frac{\tau_{t}(\ell)}{\tau_{t-1}(k)} \times P(S_t=\ell \,|\, S_{t-1}=k; \, \hat{\bm{\theta}})	 \qquad \text{if} \quad k \in \sigma_{t-1}, \, \ell \in \sigma_t\\
		& \quad  0  \qquad \qquad \qquad \qquad \qquad \qquad \qquad \quad \text{otherwise}
	\end{aligned}
	\right. \nonumber
\end{align}

\subsection{First backward step}
\label{bfb.first.backward.step}
The first backward step computes probabilities $\tau_t(s)$, the probabilities of the remaining possible states given that state $s \in \{1, \dots, K\}$ is observed at time-step $t \in \{1, \dots, T\}$: $\tau_t(s) = P(\sigma_{t+1}, \dots, \sigma_{T} \,|\, S_t=s, \hat{\bm{\theta}}) = P(S_{t+1} \in \sigma_{t+1}, \dots, S_T \in \sigma_T \,|\, S_t=s, \hat{\bm{\theta}})$. This set of probabilities is computed recursively as follows:
\begin{equation}
\label{first.backward.step}
\left\{ \begin{aligned}
    \tau_T(s) & := 1  \\ 
    \tau_t(s) &= \sum_{i \in \sigma_{t+1}} \tau_{t+1}(i) \, P(S_{t+1}=i \,|\, S_t=s; \, \hat{\bm{\theta}}).
\end{aligned}
\right.
\end{equation}
\\
\paragraph{\textbf{Proof.}}
\textit{Base case}: $t = T-1$\\
By applying the definition of $\tau_{T-1}$, we obtain:
\begin{align}
	\tau_{T-1}(s) 
	&= P(\sigma_{T} \,|\, S_{T-1}=s, \, \hat{\bm{\theta}}) = P(S_T \in \sigma_T \,|\, S_{T-1}=s; \, \hat{\bm{\theta}})\\
	&= \sum_{i \in \sigma_T} \tau_T(i) \, P(S_T=i \,|\, S_{T-1}=s; \, \hat{\bm{\theta}}).
\end{align}
\noindent \textit{Recursive case}: $t = T-2, \dots, 1$ \\
We first use the law of total probabilities (Eq. \ref{proof.tau.2}), followed by Bayes' rule (Eq. \ref{proof.tau.3}). Note that in Eq. \ref{proof.tau.3}, the probability $P(\sigma_{t+1}, \dots, \sigma_{T}\,|\,  S_{t+1}=i , S_t=s, \hat{\bm{\theta}})$ is null for $i \notin \sigma_{t+1}$ (since $\sigma_{t+1}$ is the set of possible states at time-step $t+1$); otherwise it equals $P(\sigma_{t+2}, \dots, \sigma_{T}\,|\,  S_{t+1}=i, \hat{\bm{\theta}}) = \tau_{t+1}(i)$ (Eq. \ref{proof.tau.4}). Thus, we obtain the recursive formula presented in Eq. \ref{first.backward.step}.
\begin{align}
	\tau_t(s) 
	&= P(\sigma_{t+1}, \dots, \sigma_{T} \,|\, S_t=s, \hat{\bm{\theta}}) \nonumber  \\
\label{proof.tau.2}
	&= \sum_{i=1}^K P(\sigma_{t+1}, \dots, \sigma_{T}, S_{t+1}=i \,|\, S_t=s, \hat{\bm{\theta}}) \\
\label{proof.tau.3}
	&= \sum_{i=1}^K P(\sigma_{t+1}, \dots, \sigma_{T}\,|\,  S_{t+1}=i , S_t=s, \hat{\bm{\theta}}) \, P( S_{t+1}=i \,|\, S_t=s, \hat{\bm{\theta}}) \\
\label{proof.tau.4}
	&= \sum_{i \in \sigma_{t+1}} P(\sigma_{t+2}, \dots, \sigma_{T}\,|\,  S_{t+1}=i, \hat{\bm{\theta}}) \, P( S_{t+1}=i \,|\, S_t=s, \hat{\bm{\theta}}) \\
\label{proof.tau.5}
	&= \sum_{i \in \sigma_{t+1}} \tau_{t+1}(i) \, P( S_{t+1}=i \,|\, S_t=s, \hat{\bm{\theta}}) 
\end{align}

\subsection{Forward step}
\label{bfb.forward.step}
This step allows to compute the probabilities of being in regime $s$ at time-step $t$ while observing sequence $x_1, \dots, x_t$. These probabilities, denoted by $\alpha_t(s)$, are defined as $\alpha_t(s) = P(S_t=s, X_1^t=x_1^t \,|\, X_{1-p}^0, \Sigma; \,\hat{\bm{\theta}})$ for $1 \le t \le T$, $1 \le s \le K$. They are computed as follows:
\begin{equation}
\label{forward.step}
\left\{ \begin{aligned}
	\alpha_1(s) &=  P(X_1=x_1 \,|\, X_{1-p}^{0}, S_1=s; \, \hat{\bm{\theta}}) \, P(S_1=s; \, \hat{\bm{\theta}}) \frac{\tau_1(s)}{\sum_{i \in \sigma_1} \tau_1(i) \, P(S_1=i; \, \hat{\bm{\theta}})} \\
    \alpha_t(s)  &= P(X_t=x_t \,| \, X_{t-p}^{t-1}, S_t=s; \,\hat{\bm{\theta}})  \sum_{i \in \sigma_{t-1}}  \alpha_{t-1}(i) \, P(S_t=s \,|\, S_{t-1}=i; \, \hat{\bm{\theta}}) \frac{\tau_t(s)}{\tau_{t-1}(i)} \times \bm{1}_{\{s \in \sigma_t\}}
\end{aligned}
\right.
\end{equation}
\noindent To note, the likelihood of sequence $x_1^T$ can be easily computed by integrating out $S_t$ in $\alpha_T$:
\begin{equation}
\label{bfb.comp.ll}
	P(X_1^T=x_1^T \,|\, X_{1-p}^0; \, \hat{\bm{\theta}}) = \sum_{s=1}^K \alpha_T(s).
\end{equation}
The likelihood of $N$ independent sequences is therefore calculed by multiplying the individual likelihoods across the sequences.

\paragraph{\textbf{Proof.}}
\textit{Base case:} $t=1$ \\
In Eq. \ref{proof.alpha.1}, using the conditional independence graph of the PHMC-LAR model, we transform the joint probability into two conditional probabilities, $P(X_1=x_1 \,|\, S_1=s, X_{1-p}^0; \,\hat{\bm{\theta}})$ and $P(S_1=s \,|\, \Sigma; \,\hat{\bm{\theta}})$. Then, in Eq. \ref{proof.alpha.2}, Bayes' rule is applied to the latter conditional probability. It can be easily shown that $P(\Sigma; \, \hat{\bm{\theta}}) = \sum_{i \in \sigma_1} \tau_1(i) \, P(S_1=i; \hat{\bm{\theta}})$. Thus we obtain Eq. \ref{proof.alpha.4}.
\begin{align}
	\alpha_1(s) 
	&=  P(S_1=s, X_1=x_1 \,|\, X_{1-p}^0, \Sigma; \,\hat{\bm{\theta}}) \nonumber \\
\label{proof.alpha.1}
	&= P(X_1=x_1 \,|\, S_1=s, X_{1-p}^0; \,\hat{\bm{\theta}}) \times P(S_1=s \,|\, \Sigma; \,\hat{\bm{\theta}}) \\
\label{proof.alpha.2}
	&= P(X_1=x_1 \,|\, S_1=s, X_{1-p}^0; \,\hat{\bm{\theta}}) \times \frac{P(\sigma_1, \dots, \sigma_T \,|\, S_1=s\,; \, \hat{\bm{\theta}}) \, P(S_1=s; \, \hat{\bm{\theta}})}{P(\Sigma;\,\hat{\bm{\theta}})}  \\
\label{proof.alpha.4}
	&= P(X_1=x_1 \,|\, S_1=s, X_{1-p}^0; \,\hat{\bm{\theta}}) \, P(S_1=s; \, \hat{\bm{\theta}}) \frac{\tau_1(s)}{\sum_{i \in \sigma_1} \tau_1(i) \, P(S_1=i; \hat{\bm{\theta}})}. 
\end{align}
\noindent \textit{Recursive case:} $t=2, \dots, T$ \\
As previously, the joint probability is split into two conditional probabilities (Eq. \ref{proof.alpha.1.r}). We use the law of total probabilities to introduce $S_{t-1}$ in Eq. \ref{proof.alpha.2.r}. From Eq. \ref{proof.alpha.2.r} to Eq. \ref{proof.alpha.3.r}, Bayes' rule is applied on the terms within the sum. Then, in Eq. \ref{proof.alpha.4.r}, recursive terms $\alpha_{t-1}$ weighted by probabilities $P(S_t=s \,|\, S_{t-1}=i, \Sigma; \,\hat{\bm{\theta}})$ appear within the sum. Finally, probabilities $P(S_t=s \,|\, S_{t-1}=i, \Sigma; \,\hat{\bm{\theta}})$ are computed through the calculations presented in Eq. \ref{proof.alpha.5.r}-\ref{proof.alpha.8.r}. Thus, by substituting Eq. \ref{proof.alpha.8.r} in Eq. \ref{proof.alpha.4.r}, we obtain the recursive case (Eq. \ref{forward.step}).

\begin{align}
	\alpha_t(s) 
	&= P(S_t=s, X_1^t=x_1^t \,|\, X_{1-p}^0, \Sigma; \,\hat{\bm{\theta}})  \nonumber \\
\label{proof.alpha.1.r} 
	&= P(X_t=x_t \,|\, X_1^{t-1}, S_t=s, X_{1-p}^0, \Sigma; \,\hat{\bm{\theta}}) \times P( X_1^{t-1}=x_1^{t-1}, S_t=s \,|\, X_{1-p}^0, \Sigma; \,\hat{\bm{\theta}}) \\
\nonumber 
 	&= P(X_t=x_t \,|\, X_{t-p}^{t-1}, S_t=s; \,\hat{\bm{\theta}})\\
\label{proof.alpha.2.r}
        & \quad \sum_{i=1}^K P( X_1^{t-1}=x_1^{t-1}, S_t=s, S_{t-1}=i \,|\, X_{1-p}^0, \Sigma; \,\hat{\bm{\theta}}) \\
\nonumber 
 	&= P(X_t=x_t \,|\, X_{t-p}^{t-1}, S_t=s; \,\hat{\bm{\theta}})\\
\label{proof.alpha.3.r}
        & \quad \sum_{i=1}^K P( X_1^{t-1}=x_1^{t-1}, S_{t-1}=i \,|\, X_{1-p}^0, \Sigma; \,\hat{\bm{\theta}})  P(S_t=s \,|\, S_{t-1}=i, \Sigma; \,\hat{\bm{\theta}})\\
\label{proof.alpha.4.r}
 	&= P(X_t=x_t \,|\, X_{t-p}^{t-1}, S_t=s; \,\hat{\bm{\theta}}) \sum_{i=1}^K \alpha_{t-1}(i) \, P(S_t=s \,|\, S_{t-1}=i, \Sigma; \,\hat{\bm{\theta}})
\end{align}
where
\begin{align}
\label{proof.alpha.5.r} 
	P(S_t=s \,|\, S_{t-1}=i, \Sigma; \,\hat{\bm{\theta}} )
	&= \frac{P(S_t=s, S_{t-1}=i, \Sigma; \,\hat{\bm{\theta}})}{P(S_{t-1}=i, \Sigma; \,\hat{\bm{\theta}})} \\
	&= \frac{ P(\sigma_{t+1}, \dots, \sigma_T \,|\, S_t=s, S_{t-1}=i, \sigma_1^t; \,\hat{\bm{\theta}})}{ P(\sigma_t, \dots, \sigma_T \,|\, S_{t-1}=i, \sigma_1^{t-1}; \,  \hat{\bm{\theta}}) \, P(S_{t-1}=i, \sigma_1^{t-1}; \,  \hat{\bm{\theta}})}   \nonumber \\
\label{proof.alpha.6.r} 
	& \quad \times P(S_t=s, \sigma_t \,|\, S_{t-1}=i, \sigma_1^{t-1}; \,\hat{\bm{\theta}}) \, P(S_{t-1}=i, \sigma_1^{t-1}; \,\hat{\bm{\theta}}) \\
\label{proof.alpha.7.r} 
	&= \frac{\tau_t(s)}{\tau_{t-1}(i)} \times P(S_t=s, \sigma_t \,|\, S_{t-1}=i, \sigma_{t-1}; \,\hat{\bm{\theta}}) \\
\label{proof.alpha.8.r} 
	&= \frac{\tau_t(s)}{\tau_{t-1}(i)} \times 
	\left\{ \begin{aligned}
		& P(S_t=s \,|\, S_{t-1}=i; \,\hat{\bm{\theta}}) \quad \text{if} \quad i \in \sigma_{t-1}, s \in \sigma_t \\
		& 0  \qquad \qquad \qquad  \qquad \qquad \text{otherwise.}
			\end{aligned}
	\right. 
\end{align}

\subsection{Second backward step}
\label{bfb.second.backward.step}
In this second backward step, quantities $\beta_t(s) = P(X_{t+1}^T=x_{t+1}^T \,|\, X_{t+1-p}^t, S_t = s, \Sigma; \, \hat{\bm{\theta}})$ are computed. $\beta_t(s)$ denotes the probability to observe sequence $x_{t+1}, \dots, x_T$ given that state $s$ has been observed at time-step $t$. These probabilities are recursively computed as follows:
\begin{equation}
\label{second.backward.step}
\left\{ \begin{aligned}
    \beta_T(s) &:= 1\\ 
    \beta_t(s) &= \sum_{i\in \sigma_{t+1}} \beta_{t+1}(i) \, P(S_{t+1}=i \,|\, S_t=s; \, \hat{\bm{\theta}}) \, \frac{\tau_{t+1}(i)}{\tau_t(s)}\\
               & \qquad \qquad P(X_{t+1}=x_{t+1} \,|\, X_{t+1-p}^{t}, S_{t+1}=i; \,\hat{\bm{\theta}}) \times \bm{1}_{\{s \in \sigma_t\}}. 
\end{aligned}
\right.
\end{equation}
\paragraph{\textbf{Proof.}} 
\textit{Base case:} $t=T-1$ \\
Equation \ref{proof.beta.1} is obtained by applying the law of total probabilities. In Eq. \ref{proof.beta.2}, Bayes' rule is applied to $P(S_T=i \,|\, S_{T-1} = s, \Sigma; \, \hat{\bm{\theta}})$ and a quotient of probabilities appears. Then, the numerator and denominator of this quotient are transformed into products of conditional probabilities (Eq. \ref{proof.beta.3}). In Eq. \ref{proof.beta.4}, we introduce backward propagation terms $\beta_T(i)$ and $\tau_T(i)$, which each equal one (by definition); thanks to Markov property, probability $P(S_T=i, \sigma_T \,|\, S_{T-1} = s, \sigma_1, \dots, \sigma_{T-1}; \, \hat{\bm{\theta}})$ is equal to $P(S_T=i \,|\, S_{T-1} = s; \, \hat{\bm{\theta}})$ if $i \in \sigma_T$ and $s \in \sigma_{T-1}$, and this probability is null if $i \notin \sigma_T$ and is undefined if $s \notin \sigma_{T-1}$ (hence the indicator function $\bm{1}_{\{s \, \in \sigma_{T-1}, \, i \, \in \sigma_{T}\}}$). Besides, in Eq. \ref{proof.beta.3}, a common term appears at numerator and denominator, which entails a simplification. Finally, probability $P(\sigma_T \,|\, S_{T-1} = s, \sigma_1, \dots, \sigma_{T-1}; \, \hat{\bm{\theta}})$ appearing at denominator equals $\tau_{T-1}(s)$ thanks to Markov property.\\
%
%
\begin{align}
 	\beta_{T-1}(s) 
 	&= P(X_T=x_T \,|\, X_{T-p}^{T-1}, S_{T-1} = s, \Sigma; \, \hat{\bm{\theta}}) \nonumber \\
\label{proof.beta.1}
	&= \sum_{i=1}^K P(X_T=x_T \,|\, X_{T-p}^{T-1}, S_T=i, \Sigma; \, \hat{\bm{\theta}}) \, P(S_T=i \,|\, S_{T-1} = s, \Sigma; \, \hat{\bm{\theta}}) \\
\label{proof.beta.2}	
	&= \sum_{i=1}^K P(X_T=x_T \,|\, X_{T-p}^{T-1}, S_T=i; \, \hat{\bm{\theta}}) \, \frac{P(S_T=i, S_{T-1} = s, \Sigma; \, \hat{\bm{\theta}})}{P(S_{T-1} = s, \Sigma; \, \hat{\bm{\theta}})}  \\
	&= \sum_{i=1}^K P(X_T=x_T \,|\, X_{T-p}^{T-1}, S_T=i; \, \hat{\bm{\theta}})  \,\nonumber \\
\label{proof.beta.3}	
	&  \quad \qquad \times \frac{ P(S_T=i, \sigma_T \,|\, S_{T-1} = s, \sigma_1, \dots, \sigma_{T-1}; \, \hat{\bm{\theta}}) \, P(S_{T-1} = s, \sigma_1, \dots, \sigma_{T-1}; \, \hat{\bm{\theta}}) } { P(\sigma_T \,|\, S_{T-1} = s, \sigma_1, \dots, \sigma_{T-1}; \, \hat{\bm{\theta}}) \, P(S_{T-1} = s, \sigma_1, \dots, \sigma_{T-1}; \, \hat{\bm{\theta}})}   \\
	&= \sum_{i=1}^K \beta_{T}(i) \, P(X_T=x_T \,|\, X_{T-p}^{T-1}, S_T=i; \, \hat{\bm{\theta}}) \, \frac{\tau_T(i)}{ \tau_{T-1}(s)}  \nonumber \\
\label{proof.beta.4}
	&  \quad  \qquad \times P(S_T=i \,|\, S_{T-1} = s; \, \hat{\bm{\theta}}) \times \bm{1}_{\{s \, \in \sigma_{T-1}, \, i \, \in \sigma_{T}\}.} 
\end{align} 
\noindent \textit{Recursive case:} $t=T-2, \dots, 1$ \\
The application of the law of total probabilities yields Eq. \ref{proof.beta.1.r}. In Eq. \ref{proof.beta.2.r}, $X_{t+2}^T$ then $X_{t+1}$ are marginalized, which allows to make appear the recursive term $\beta_{t+1}$ together with the conditional probability of $X_{t+1}$ given $S_{t+1}$ and past values in Eq. \ref{proof.beta.3.r}. As in the base case, probability $P(S_{t+1} = i \,|\, S_t=s, \Sigma; \, \hat{\bm{\theta}})$ is computed using Bayes' rule (Eq. \ref{proof.beta.4.r}-\ref{proof.beta.7.r}).\\
\begin{align}
	\beta_t(s) 
	&= P(X_{t+1}^T=x_{t+1}^T \,|\, X_{t+1-p}^t, S_t = s, \Sigma; \, \hat{\bm{\theta}}) \nonumber \\
\label{proof.beta.1.r}
	&= \sum_{i=1}^K P(X_{t+1}=x_{t+1}, X_{t+2}^T=x_{t+2}^T, S_{t+1} = i \,|\, X_{t+1-p}^t, S_t=s, \Sigma; \, \hat{\bm{\theta}}) \\	
	&= \sum_{i=1}^K P(X_{t+2}^T=x_{t+2}^T \,|\, X_{t+1-p}^{t+1}, S_{t+1} = i, \Sigma; \, \hat{\bm{\theta}}) \,  \nonumber \\
\label{proof.beta.2.r}
	&  \quad \qquad \times P(X_{t+1}=x_{t+1}  \,|\, X_{t+1-p}^t, S_{t+1} = i, \Sigma; \, \hat{\bm{\theta}}) \, P(S_{t+1} = i \,|\, S_t=s, \Sigma; \, \hat{\bm{\theta}})\\
\label{proof.beta.3.r}	
	&= \sum_{i=1}^K \beta_{t+1}(i) \, P(X_{t+1}=x_{t+1}  \,|\, X_{t+1-p}^t, S_{t+1} = i, \Sigma; \, \hat{\bm{\theta}}) P(S_{t+1} = i \,|\, S_t=s, \Sigma; \, \hat{\bm{\theta}})
\end{align}
where
\begin{align}
\label{proof.beta.4.r}
	P(S_{t+1} = i \,|\, S_t=s, \Sigma; \, \hat{\bm{\theta}}) 
	&=\frac{P(S_{t+1} = i, S_t=s, \Sigma; \, \hat{\bm{\theta}})}{P(S_t=s, \Sigma; \, \hat{\bm{\theta}})} \\
	&= (\sigma_{t+2}, \dots, \sigma_T \,|\, S_{t+1} = i, S_t=s, \sigma_1, \dots, \sigma_{t+1}; \, \hat{\bm{\theta}}) \, \times \nonumber \\
	& \frac{P(S_{t+1} = i, \sigma_{t+1} \,|\, S_t=s, \sigma_1, \dots, \sigma_t; \, \hat{\bm{\theta}}) \, P(S_t=s, \sigma_1, \dots, \sigma_t; \, \hat{\bm{\theta}})}{P(\sigma_{t+1}, \dots, \sigma_T \,|\, S_t=s, \sigma_1, \dots, \sigma_t; \, \hat{\bm{\theta}}) \, P(S_t=s, \sigma_1, \dots, \sigma_t; \, \hat{\bm{\theta}})} \\
	&= \frac{ \tau_{t+1}(i) }{ \tau_t(s) } \times P(S_{t+1} = i, \sigma_{t+1} \,|\, S_t=s, \sigma_t; \, \hat{\bm{\theta}}) \\
\label{proof.beta.7.r}	
	&= \frac{ \tau_{t+1}(i) }{ \tau_t(s) } \times P(S_{t+1} = i \,|\, S_t=s; \, \hat{\bm{\theta}}) \times \bm{1}_{\{s \, \in \sigma_t, \, i \, \in \sigma_{t+1}\}}.  
\end{align}

\subsection{Scaling of backward-forward-backward algorithm}
\label{BFB.scaling}
For large sequences, {\it i.e.} large value of $T$, the  quantities $\tau_t(s)$, $\alpha_t(s)$ and $\beta_t(s)$ tend to zero as products of probabilities. Thus, the computations will require beyond the precision range of machine and PHMC-LAR parameter estimate will be inaccurate. Generally, this problem is solved by normalizing $\tau_t(s)$, $\alpha_t(s)$ and $\beta_t(s)$ by a term of same order of magnitude \citep{florez_2020_incremental_scaling-in-baum-welch-algo, koenig_simmons_1996_jour-ieee_scaling-in-baum-welch-algo}. Thus, we propose the following normalization:
\begin{align}
\label{tau.tilde}
 	\tilde{\tau}_t(s) &= \frac{\tau_t(s)}{P(\sigma_t, \dots, \sigma_T \,|\, \sigma_{t-1}; \, \bm{\hat{\theta}})}, \\
\label{alpha.tilde}
	\tilde{\alpha}_t(s) &= \frac{\alpha_t(s)}{P(X_1^t=x_1^t \,|\, X_{1-p}^0, \Sigma; \, \bm{\hat{\theta}})}, \\
\label{beta.tilde}
	\tilde{\beta}_t(s) &= \frac{\beta_t(s)}{P(X_t^T=x_t^T \,|\, X_{1-p}^{t-1}, \Sigma; \, \bm{\hat{\theta}})}.
\end{align}
As previouly, $\tilde{\tau}_t(s)$, $\tilde{\alpha}_t(s)$ and $\tilde{\beta}_t(s)$ can be computed recursively. The recursive formula for these quantities can be deduced from those of $\tau_t(s)$ (Eq. \ref{first.backward.step}), $\alpha_t(s)$ (Eq. \ref{forward.step}) and $\beta_t(s)$ (Eq. \ref{second.backward.step}). To do so, Eq. \ref{first.backward.step}, \ref{forward.step} and \ref{second.backward.step} are respectively divided by the normalization terms $P(\sigma_t, \dots, \sigma_T \,|\, \sigma_{t-1}; \, \bm{\hat{\theta}})$, $P(X_1^t=x_1^t \,|\, X_{1-p}^0, \Sigma; \, \bm{\hat{\theta}})$ and $P(X_t^T=x_t^T \,|\, X_{1-p}^{t-1}, \Sigma; \, \bm{\hat{\theta}})$. After decomposing the formula obtained and after some calculations, we obtain the subsequent recurvive formulas for $\tilde{\tau}_t$, $\tilde{\alpha}_t$ and $\tilde{\beta
}_t$.

\paragraph{First backward propagation}
\begin{align}
\label{scaled.tau}
\left\{ \begin{aligned}
    \tilde{\tau}_T(s) &=  \frac{1}{P(\sigma_T \,|\, \sigma_{T-1}; \, \hat{\bm{\theta}})}  \\
    \tilde{\tau}_t(s) &= \sum_{i \in \sigma_{t+1}} \tilde{\tau}_{t+1}(i) \, \frac{ P(S_{t+1}=i \,|\, S_t=s; \, \hat{\bm{\theta}}) } { P(\sigma_t \,|\, \sigma_{t-1}; \, \bm{\hat{\theta}})}, 
    \quad \text{for} \quad t= T-1, \dots, 1,
\end{aligned}
\right.
\end{align}
with 
\begin{equation}
\label{prob.sigma}
\begin{split}
	P(\sigma_t \,|\, \sigma_{t-1}; \, \bm{\hat{\theta}}) 
	&= P(S_t \in \sigma_t \,|\, S_{t-1} \in \sigma_{t-1}; \bm{\hat{\theta}}) =  \sum_{i \in \sigma_{t-1}} \sum_{j \in \sigma_t} P(S_t = j \,|\, S_{t-1} = i; \, \bm{\hat{\theta}}). \\
	P(\sigma_1; \, \bm{\hat{\theta}}) 
	&= P(S_1 \in \sigma_1; \bm{\hat{\theta}}) = \sum_{i \in \sigma_1} P(S_1 = i; \, \bm{\hat{\theta}}).
\end{split}
\end{equation}
\paragraph{Forward propagation}
\begin{align}
\label{scaled.alpha}
\left\{ \begin{aligned}
	\tilde{\alpha}_1(s) &=  \frac{P(X_1=x_1 \,|\, X_{1-p}^{0}, S_1=s; \, \hat{\bm{\theta}}) \, P(S_1=s; \, \hat{\bm{\theta}})}{ C_1 } \times \frac{\tilde{\tau}_1(s)}{\sum_{i \in \sigma_1} \tilde{\tau}_1(i) \, P(S_1=i; \, \hat{\bm{\theta}})} \\
    \tilde{\alpha}_t(s)  
    &= P(X_t=x_t \,| \, X_{t-p}^{t-1}, S_t=s; \,\hat{\bm{\theta}}) \left[ \sum_{i \in \sigma_{t-1}}  \tilde{\alpha}_{t-1}(i) \, P(S_t=s \,|\, S_{t-1}=i; \, \hat{\bm{\theta}}) \frac{\tilde{\tau}_t(s)}{\tilde{\tau}_{t-1}(i)} \right] \, \\
    &\quad  \times  \frac{1}{P(\sigma_{t-1} \,|\, \sigma_{t-2}; \, \hat{\bm{\theta}}) \, C_t} \times \bm{1}_{\{s \in \sigma_t\}}
\end{aligned}
\right.
\end{align}
with $P(\sigma_{t-1} \,|\, \sigma_{t-2}; \, \hat{\bm{\theta}})$ defined in Eq. \ref{prob.sigma} and $C_t$ the scaling term defined and computed as follows:
\begin{align}
\label{c.1.def}
	C_1 &= P(X_1=x_1 \,|\, X_{1-p}^0, \Sigma; \, \hat{\bm{\theta}} ) = \sum_{i \in \sigma_1} P(X_1=x_1 \,|\, X_{1-p}^{0}, S_1=i; \, \hat{\bm{\theta}}) \, P(S_1=i; \, \hat{\bm{\theta}}) \\
\label{c.t.def}
	C_t &= P(X_t=x_t \,|\, X_{1-p}^{t-1}, \Sigma; \, \hat{\bm{\theta}}) \quad \text{for} \quad t=2, \dots, T\\
		&= \sum_{s \in \sigma_t} P(X_t=x_t \,|\, X_{t-p}^{t-1}, S_t=s; \, \hat{\bm{\theta}}) 
		\times \left[ \sum_{i \in \sigma_{t-1}} \tilde{\alpha}_{t-1}(i) \, P(S_t=s \,|\, S_{t-1}=i; \, \hat{\bm{\theta}}) \right].
\end{align}
The proof is straightforward and is left to the reader. Note that 
$P(X_1^T=x_1^T \,|\, X_{1-p}^0; \, \hat{\bm{\theta}}) = \prod_{t=1}^T C_t$.
\paragraph{Second backward propagation}
\begin{align}
\label{scaled.beta}
\left\{ \begin{aligned}
    \tilde{\beta}_T(s) &= \frac{1}{C_T} \\
    \tilde{\beta}_t(s) 
    &= \sum_{i\in \sigma_{t+1}} \left[ \tilde{\beta}_{t+1}(i) \, P(S_{t+1}=i \,|\, S_t=s; \, \hat{\bm{\theta}}) \, \frac{\tilde{\tau}_{t+1}(i)}{\tilde{\tau}_t(s)} \, 
    P(X_{t+1}=x_{t+1} \,|\, X_{t+1-p}^{t}, S_{t+1}=i; \,\hat{\bm{\theta}}) \right] \\
    & \quad \times \frac{1}{P(\sigma_t \,|\, \sigma_{t-1}; \, \hat{\bm{\theta}}) \, C_t} \times \bm{1}_{\{s \in \sigma_t\}} 
\end{aligned}
\right.
\end{align}
where $C_t$ and $P(\sigma_t \,|\, \sigma_{t-1}; \, \hat{\bm{\theta}})$ are defined in Eq. \ref{c.1.def}-\ref{c.t.def} and Eq. \ref{prob.sigma} respectively.
\paragraph{$\xi_t(k, \ell)$ computation\\} 
In Eq. \ref{def.xi.app} probabilities $\xi_t(k, \ell)$ are defined in function of quantities $\tau_t$, $\tau_{t-1}$, $\alpha_{t-1}$ and $\beta_t$. These quantities can be easily expressed in function of their normalized versions $\tilde{\tau}_t$, $\tilde{\tau}_{t-1}$, $\tilde{\alpha}_{t-1}$ and $\tilde{\beta}_t$ using Eq. \ref{tau.tilde}, \ref{alpha.tilde} and \ref{beta.tilde}. After substituting $\tau_t$, $\tau_{t-1}$, $\alpha_{t-1}$ and $\beta_t$ by the resulting expressions and after some simplifications, we obtain the following formula:
\begin{align}
\label{scaled.xi}
    \xi_t(k, \ell) 
    &= \frac{ \tilde{\beta}_t(l) \, P(S_t=\ell \,|\, S_{t-1}=k; \, \hat{\bm{\theta}}) \, P(X_t=x_t \,|\, X_{t-p}^{t-1}, S_t=\ell; \, \hat{\bm{\theta}}) \, \tilde{\alpha}_{t-1}(k) \, \tilde{\tau}_t(\ell)} {P(\sigma_{t-1} \,|\, \sigma_{t-2}; \, \hat{\bm{\theta}}) \, \tilde{\tau}_{t-1}(k)}  \nonumber \\
    & \quad \times \bm{1}_{\{\ell \in \sigma_t, \, k \in \sigma_{t-1}\}}.
\end{align}

\section{Appendix: decomposition of $Q(\theta, \hat{\theta}_{n-1})$}
\label{decomposition.of.Q}
From Eq. \ref{e.step} and \ref{log.likelihood.several.seq}, it is straightforward to show that $Q(\theta, \hat{\theta}_{n-1})$ can be decomposed as the sum of quantities $Q_S(\theta^{(S)}, \hat{\theta}_{n-1})$ and $Q_X(\theta^{(X)}, \hat{\theta}_{n-1})$:
%
%
\begin{equation}
\begin{split}
 Q_S(\theta^{(S)}, \hat{\theta}_{n-1}) &= \sum_{i=1}^{N} \, \sum_{s \in \mathbf{K}} \ln\left(P(S_1^{(i)}=s; \, \hat{\theta}_{n-1}) \right) \, P(S_1^{(i)}=s \,|\, [X^{(i)}]_{1-p}^{T_i}, \Sigma^{(i)}; \, \hat{\theta}_{n-1}) \\
    &+ \sum_{i=1}^{N} \, \sum_{t=2}^{T_i} \, \sum_{(s, s') \in \mathbf{K}^2} \ln\left(P(S_t^{(i)}=s \,|\, S_{t-1}^{(i)}=s'; \, \theta^{(S)}) \right)\\
    & \qquad \qquad \qquad \qquad \quad P(S_t^{(i)}=s, S_{t-1}^{(i)}=s' \,|\, [X^{(i)}]_{1-p}^{T_i}, \Sigma^{(i)}; \, \hat{\theta}_{n-1}),
\end{split}
\end{equation}
\begin{equation}
\begin{split}
Q_X(\theta^{(X)}, \hat{\theta}_{n-1}) &= \sum_{i=1}^{N} \, \sum_{t=1}^{T_i} \, \sum_{s \in \mathbf{K}} \ln \left( P (x_t^{(i)} \,|\, X_{t-1}^{(i)}, ..., X_{t-p}^{(i)}, S_t^{(i)}=s; \, \theta^{(X)})\right)\\
  & \qquad \qquad \qquad \quad P(S_t^{(i)}=s \,|\, [X^{(i)}]_{1-p}^{T_i}, \Sigma^{(i)}; \, \hat{\theta}_{n-1}),
\end{split}
\end{equation}
with $\mathbf{K}=\{1, \dots, K\}$.

\bibliographystyle{spbasic}      
\bibliography{references}   

%


\end{document}